\pdfoutput=1

\documentclass[11pt]{article}

\usepackage[final]{acl}

\usepackage{times}
\usepackage{latexsym}

\usepackage[T1]{fontenc}

\usepackage[utf8]{inputenc}

\usepackage{microtype}

\usepackage{inconsolata}

\usepackage{graphicx}

\usepackage{amsmath}
\usepackage{algorithm}
\usepackage{algpseudocode}
\usepackage{listings}

\lstdefinestyle{mystyle}{
    columns=fixed,
    basewidth=0.5em,
    basicstyle=\ttfamily
}

\title{Spider4SSC \& S2CLite: A text-to-multi-query-language dataset using lightweight ontology-agnostic SPARQL to Cypher parser}

\author{Martin Vejvar \\
  Graduate School of Engineering Science \\
  Yokohama National University \\
  \texttt{vejvar-martin-km@ynu.jp} \\\And
  Yasutaka Fujimoto \\
  Faculty of Engineering \\
  Yokohama National University \\
  \texttt{fujimoto@ynu.ac.jp} \\}

\begin{document}
\maketitle
\begin{abstract}
We present Spider4SSC dataset and S2CLite parsing tool. S2CLite is a lightweight, ontology-agnostic parser that translates SPARQL queries into Cypher queries, enabling both in-situ and large-scale SPARQL to Cypher translation. Unlike existing solutions, S2CLite is purely rule-based (inspired by traditional programming language compilers) and operates without requiring an RDF graph or external tools. Experiments conducted on the BSBM42 and Spider4SPARQL datasets show that S2CLite significantly reduces query parsing errors, achieving a total parsing accuracy of 77.8\% on Spider4SPARQL compared to 44.2\% by the state-of-the-art S2CTrans. Furthermore, S2CLite achieved a 96.6\% execution accuracy on the intersecting subset of queries parsed by both parsers, outperforming S2CTrans by 7.3\%. We further use S2CLite to parse Spider4SPARQL queries to Cypher and generate Spider4SSC, a unified Text-to-Query language (SQL, SPARQL, Cypher) dataset with 4525 unique questions and 3 equivalent sets of 2581 matching queries (SQL, SPARQL and Cypher). We open-source S2CLite for further development on GitHub\footnote{https://github.com/vejvarm/S2CLite} and provide the clean Spider4SSC dataset for download\footnote{\url{https://www.dropbox.com/scl/fi/v4wa82hpe2s5p5uh4bgnj/Spider4SSC.zip?rlkey=rb447tkhjq38ugnd0v8rpt2bh&e=3&st=s94sdngb}}.
\end{abstract}

\section{Introduction}
Knowledge graphs are well-established structured data models where nodes represent unique entities and oriented edges denote their properties and interactions. In the knowledge graph community, the predominant model is Resource Description Framework (RDF) \cite{cyganiak2014rdf}, as its SPARQL query language follows highly defined W3C standards. However, in the pursuit of efficiency, labelled property graphs have recently gained more traction in industry, with Neo4j and Cypher being the most prominent property graph store and query language respectively \cite{francis2018cypher}. Unlike RDF graph, in which every resource, relationship or property is represented as a triple, property graph model defines resource nodes and edges as objects with properties, leading to compact and intuitive representation of structured data \cite{szarnyas2018propertygraphs} and generally reduced query execution times \cite{zhao2023s2ctrans}. This pressing need for deeper interconnection and representation between the two storage systems and languages has been addressed by several approaches. \citet{barrasa2021neosemantics} develop Neosemantics, a universal strategy for converting an RDF graph into an equivalent property graph. \citet{liu2021unikg} circumvent the languages entirely, by presenting a unified storage scheme, which both Cypher and SPARQL can operate on. On the language side, \citet{a2022sparql2cypher} and \citet{zhao2023s2ctrans} leverage Jena ARQ \cite{wilkinson2006jenaarq} and Neosemantics to develop pipelines for parsing SPARQL queries into equivalent Cypher queries. While robust in nature, their reliance on external tools and insufficient support for common SPARQL query types lead us to present S2CLite, a lightweight and purely algorithmic SPARQL to Cypher parser. 

\paragraph{S2CLite} operates out of the box, requiring neither a specific database file nor a connection to any data store system. In \S\ref{sec:related-work} we discuss the significance of S2CLite in the context of existing research. In \S\ref{sec:methods}, we describe S2CLite functionality in detail. In \S\ref{sec:experiments} we compare our results to \citeposs{zhao2023s2ctrans} S2CTrans parser (current SotA) and evaluate both parsers with respect to the error rate of unparsed SPARQL queries (\S\ref{subsubsec:exp-parsing-errors}) and execution accuracy of parsed queries (\S\ref{subsubsec:exp-exec-acc}) for BSBM \cite{bizer2008bsbm} and Spider4SPARQL \cite{kosten2023spider4sparql} datasets. In \S\ref{sec:conclusion} we further discuss the experiment results and conclude our findings. In \S\ref{sec:limitations}, we present limitations of S2CLite and future work.


\section{Contributions} 
\subsection{S2CLite parser}
We propose \textbf{S2CLite} for rule-based parsing of SPARQL queries to Cypher. Namely, we design the algorithms for Visitor (\S\ref{subsec:visitor}) and Interpreter (\S\ref{subsec:interpreter}) classes presented in \S\ref{sec:methods} below.
\subsection{Spider4SSC dataset}
Using S2CLite, we parse Spider4SPARQL queries into matching Cypher counterparts and compose \textbf{Spider4SSC}, an aggregate Text-to-Query language dataset with \textbf{4525 unique} (question, sql, sparql, cypher) \textbf{entries} where SQL, SPARQL and Cypher are semantically equivalent queries operating over \textbf{159 executable} SQLite/RDF graph databases. We collect the SQL queries from Spider \cite{yu-etal-2018-spider} and SPARQL queries from Spider4SPARQL \cite{kosten2023spider4sparql}. See Listing \ref{lst:s4ssc-example} for one sample entry from Spider4SSC dev set and refer to \S\ref{app:s4ssc} for dataset creation details and download.

\section{Related Work}
\label{sec:related-work}
Due to the semantic differences between Cypher and SPARQL, developing a universal standard for their interoperability poses significant challenges \cite{liu2021unikg}.

\subsection{SPARQL to Cypher Conversion Tools}
\label{subsec:s2c-parsers}
This section reviews existing SPARQL to Cypher conversion tools and highlights the key limitations they face.

\subsubsection{S2C transpiler} 
Developed by \citet{a2022sparql2cypher}, SPARQL to Cypher transpiler uses Apache Jena ARQ \cite{wilkinson2006jenaarq} to generate an Abstract Syntax Tree (AST), which is then traversed by their custom Visitor class\footnote{\url{https://github.com/kracr/sparql-cypher-transpiler}}. S2C transpiler supports only basic graph patterns, limiting its applicability to queries without any solution modifiers. Due to this limitation, we do not consider it in our experimental section (\S\ref{sec:experiments}).

\subsubsection{S2CTrans} 
\label{subsubsec:s2ctrans}
Introduced more recently by \citet{zhao2023s2ctrans}, S2CTrans demonstrates a relational algebra-compliant semantic equivalence between SPARQL and Cypher graph queries. S2CTrans pipeline is robust, leveraging Jena ARQ, Neo4j Neosemantics \cite{barrasa2021neosemantics} and Cypher DSL \cite{meier2021cypherdsl} in tandem with custom Pattern Matching and Solution Modifier modules to produce syntactically and semantically equivalent Cypher queries. However, this approach presents significant limitations:
\begin{enumerate}
    \item \textbf{Reliance on RDF graph:} S2CTrans requires the RDF graph data file and a connection to Neo4j's Neosemantics. Specifically, S2CTrans loads RDF data via Neosemantics to access the ontology of node properties and relationships. This dependency makes S2CTrans unsuitable for standalone SPARQL to Cypher conversion, as it necessitates the availability of the corresponding RDF knowledge graph.
    \item \textbf{Limited query types:} S2CTrans supports a substantial amount of SPARQL query types, including complex property paths (\lstinline{subj p1/p2/p3 obj}) and optional patterns (OPTIONAL), but lacks support for other fundamental syntax, such as solution grouping (GROUP BY) or projection of all bound variables (SELECT *).
\end{enumerate}
\textbf{S2CLite} addresses the above limitations with a purely algorithmic solution that operates independently of any ontology or knowledge graph.



\subsection{Datasets}
To asses and compare S2CLite with existing work, we use the two datasets below. 
\subsubsection{BSBM42}
\label{subsubsec:ds-bsbm42}
Originally presented by \citet{bizer2008bsbm}, the Berlin SPARQL Benchmark (BSBM) was designed for evaluating the performance of different RDF graph storage systems on large and complex graph. \textbf{BSBM42} contains 42 unique queries (executable on the BSBM knowledge graph) designed by \citet{zhao2023s2ctrans} to test their S2CTrans parser. In \S\ref{subsec:experiments-bsbm42}, we evaluate S2CLite on the same set of queries.

\subsubsection{Spider4SPARQL}
\label{subsubsec:ds-s4s}
Spider dataset is a large scale Text-to-SQL composite of 200 SQLite relational databases and 5693 unique SQL queries, paired with 10181 natural language questions \cite{yu-etal-2018-spider}. \textbf{Spider4SPARQL} is a sibling dataset aimed at Text-to-SPARQL. \citet{kosten2023spider4sparql} materialised a large portion of the Spider SQLite relational databases (166 in total) into RDF graphs using the Ontop \cite{calvanese2017ontop} Virtual Knowledge Graph system. They further used SemQL, a context-free grammar, as an intermediary language between SQL and SPARQL to facilitate their translation, producing 4721 SPARQL queries, equivalent in functionality to their respective SQL queries. In \S\ref{subsec:experiments-s4s}, we use Spider4SPARQL to evaluate both S2CTrans and S2CLite with respect to parsing errors and execution accuracy.

\section{Methods}
\label{sec:methods}
\begin{figure}[htb]
  \centering
  \includegraphics[width=.8\columnwidth]{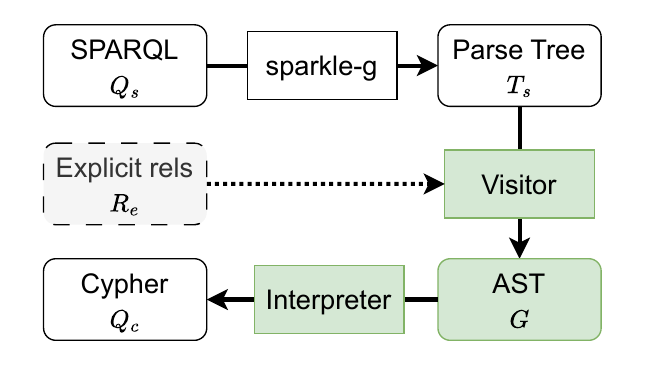}
  \caption{Overview of S2CLite parser. SPARQL query $Q_s$ is transformed into SPARQL 1.1 Parse tree $T_s$ using \citet{2013sparkleg} grammar. Visitor traverses $T_s$ and generates a Cypher pattern abstract syntax tree ($\mathcal{G}$), which is then used by Interpreter to build the final Cypher query $Q_c$. Visitor can optionally take a set of explicit relationship type names ($R_e$), enforcing them as relationships instead of node properties (see \S\ref{app:explcit-rel-type-names}).}
  \label{fig:s2clite-pipeline}
\end{figure}

As previously stated, S2CLite is an algorithmic SPARQL to Cypher parser. Given a SPARQL query, S2CLite generates a semantically equivalent Cypher query. The complete parsing pipeline is depicted in Figure \ref{fig:s2clite-pipeline}. Our approach takes inspiration from traditional programming language compilers, where initially a parse tree is generated from the query to catch any syntax errors, followed by a walker class (visitor or listener), extracting relevant information from the parse tree, which is then interpreted into the output language. 

\begin{figure*}[htb]
  \centering
  \includegraphics[width=.95\textwidth]{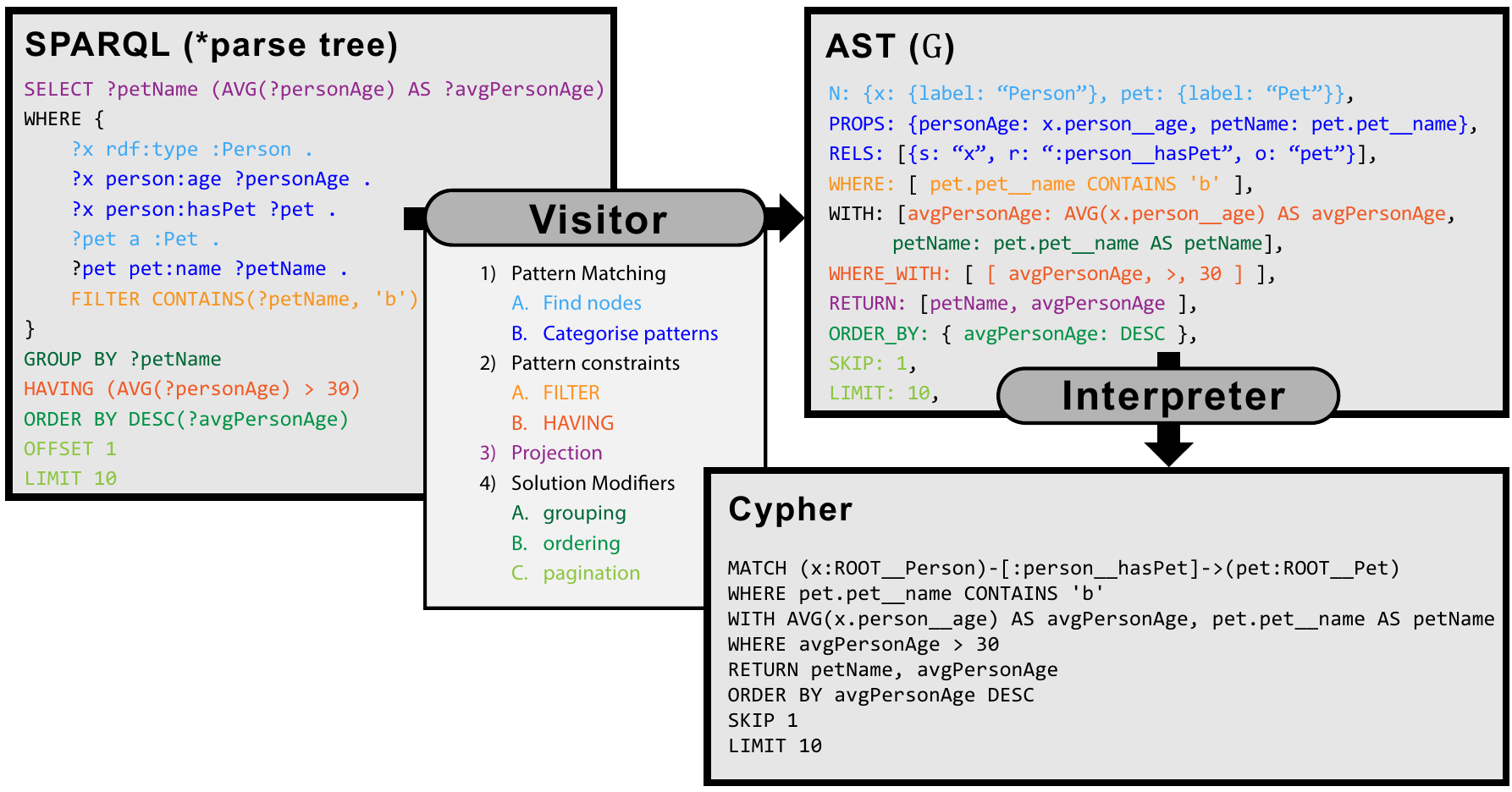}
  \caption{Step by step example of Visitor constructing a simplified version of AST ($\mathcal{G}$) from given SPARQL query and the final Cypher query assembled by the Interpreter. Parse tree creation step is omitted for terseness (see \S\ref{app:sparkle-g}). Steps of the Visitor are colour-coded with their respective visited SPARQL pattern and corresponding generated AST entry (see \S\ref{app:ast-example} for full $\mathcal{G}$). The SPARQL query can be verbalised as: \textit{"Average age of people who own pets with the same name containing the letter 'b', filtered and ordered by owner age and limited to top 10, skipping first one."}}
  \label{fig:visitor-overview}
\end{figure*}

\paragraph{0. Parse tree ($T_s$)} of a given SPARQL query is generated using sparkle-g (see \S\ref{app:sparkle-g}). This guarantees that the query conforms to SPARQL 1.1 syntax. See Fig. \ref{fig:sparql-query-01-parse-tree} for parse tree example.

\paragraph{1. Visitor} walks the nodes of $T_s$ and maps visited SPARQL patterns (SELECT, WHERE, etc.) to their respective Cypher equivalents (RETURN, MATCH, resp.), generating an Abstract Syntax Tree (AST, $\mathcal{G}$). 

\paragraph{2. Interpreter} builds the final Cypher query from the patterns and solution modifiers in $\mathcal{G}$. 

\subsection{SPARQL vs. Cypher}
\label{sec:sparql-vs-cypher}
Before detailing the Visitor and Interpreter functionality, it is prudent to  discuss the general mapping of SPARQL language (RDF graph) to Cypher language (property graph). We will only focus on the necessities for this paper. For in-depth mapping of Cypher using relational algebra, refer to \cite{marton2017opencypher, zhao2023s2ctrans}.

\begin{table}[htb]
\centering
\caption{Overview of mapping SPARQL patterns and solution modifiers to their Cypher equivalent where $e_i$ stands for any expression (variable, literal or function) and $\mathcal{C}$ marks any constraint, such as $a < b$ or $a$ \lstinline{IN} $b$ etc.}
\label{tab:sparql-to-cypher-map}
\resizebox{\columnwidth}{!}{%
\begin{tabular}{|l|l|}
\hline
\multicolumn{1}{|c|}{\textbf{SPARQL}}                                                                    & \multicolumn{1}{c|}{\textbf{Cypher}}                                                                                  \\ \hline
SELECT ($e_1$ AS ?x) ?y                                                                                   & RETURN $e_1$ AS x, y                                                                                                  \\ \hline
WHERE \{ ?x a :L . ?x :rel ?y .\}                                                                        & MATCH (x:L)-{[}:rel{]}-\textgreater{}(y)                                                                              \\ \hline
FILTER $\mathcal{C}(e_1, [e_2])$                                                                                   & WHERE $\mathcal{C}(e_1, [e_2])$                                                                                                 \\ \hline
\begin{tabular}[c]{@{}l@{}}SELECT $e_1$ AS ?x, {[}$e_2$ AS ?y{]}\\ HAVING $\mathcal{C}$(?x, {[}?y{]})\end{tabular} & \begin{tabular}[c]{@{}l@{}}WITH $e_1$ AS x, {[}$e_2$ AS y{]}\\ WHERE $\mathcal{C}$(x, {[}y{]})\\ RETURN x, {[}y{]}\end{tabular} \\ \hline
\begin{tabular}[c]{@{}l@{}}SELECT ?x (AVG(?y) AS ?avg)\\ GROUP BY ?x\end{tabular}                       & \begin{tabular}[c]{@{}l@{}}WITH x, AVG(y) AS avg\\ RETURN x, avg\end{tabular}                                       \\ \hline
ORDER BY ASC(?x) DESC(?y)                                                                                & ORDER BY x ASC, y DESC                                                                                                \\ \hline
LIMIT n                                                                                                  & LIMIT n                                                                                                               \\ \hline
OFFSET n                                                                                                 & SKIP n                                                                                                                \\ \hline
\end{tabular}%
}
\end{table}

\paragraph{Mapping SPARQL to Cypher:} In Table \ref{tab:sparql-to-cypher-map} we show a general mapping of SPARQL base clauses to their Cypher equivalent. Below are the main syntactic discrepancies between the languages.

\paragraph{Triple patterns} in WHERE block of SPARQL map to MATCH in Cypher. This is directly related to the basic pattern in the respective graphs. While in RDF graphs, the most atomic pattern is represented by a single triple (subject, predicate, object), in property graphs, every graph node and edge is an object. Objects can have zero or more types (labels) and have zero or more named properties assigned to them. Consider the SPARQL and parsed Cypher queries in Figure \ref{fig:visitor-overview} as example.




\paragraph{WITH clause} is a unique Cypher clause, which can be used to project and manipulate results at any point in the Cypher query. WITH has many uses\footnote{\url{https://neo4j.com/docs/cypher-manual/current/clauses/with/}}, from which we stress the ability to group results and apply aggregate functions to variables. For instance, '\lstinline{WITH x, AVG(y) AS avg}' in Tab. \ref{tab:s2c-pattern-matching} projects the averages of $\texttt{y}$ (as $\texttt{avg}$), grouped by $\texttt{x}$.

\subsection{Visitor}
\label{subsec:visitor}
Our Visitor is a Python implementation of a Parse tree walker class, sub-classed from ANTLR4 ParseTreeVisitor\footnote{\url{https://www.antlr.org/api/Java/org/antlr/v4/runtime/tree/ParseTreeVisitor.html}}. Its purpose is to traverse the given parse tree ($T_s$) based on defined logic. At each node of $T_s$, Visitor can add some relevant information to the AST ($\mathcal{G}$, represented as Python dictionary) and decide which child nodes to visit next.

\paragraph{Building the AST ($\mathcal{G}$):} First, we initialise $\mathcal{G}$ with empty fields using alg. in \S\ref{alg:init}. Subsequently, Figure \ref{fig:visitor-overview} shows the general steps Visitor takes as it builds $\mathcal{G}$ from a given parse tree. We elaborate on the step by step process below.

\subsubsection{Pattern Matching}
\label{subsubsec:pattern-matching}
Given a WHERE block with triple patterns $(s \ p \ o) \in \mathcal{T}$, we can divide them into 4 types (see Tab. \ref{tab:s2c-pattern-matching}). 

\begin{table}[ht]
\centering
\caption{Types of SPARQL triple patterns and their respective mapping to equivalent Cypher constructs.}
\label{tab:s2c-pattern-matching}
\resizebox{.48\textwidth}{!}{%
\begin{tabular}{|l|l|l|}
\hline
\multicolumn{1}{|c|}{\textbf{pattern type}} & \multicolumn{1}{c|}{\textbf{triple pattern}}                                                                         & \multicolumn{1}{c|}{\textbf{cypher pattern}}                                                                 \\ \hline
\textbf{1) class label}                     & \begin{tabular}[c]{@{}l@{}}$(?x \ \texttt{rdf:type} \ \mathcal{L})$\\ $(?y \ \texttt{a} \ \mathcal{L})$\end{tabular} & \begin{tabular}[c]{@{}l@{}}MATCH ($x$:$\mathcal{L}$)\\ MATCH ($y$:$\mathcal{L}$)\end{tabular} \\ \hline
\textbf{2) value constraint}                & $(?x \ \texttt{:name} \ \texttt{'Emma'})$                                                                            & MATCH ($x$ \{name: 'Emma'\})                                                                                 \\ \hline
\textbf{3) variable property}               & $(?x \ \texttt{:age} \ ?y)$                                                                                          & \begin{tabular}[c]{@{}l@{}}MATCH ($x$)\\ RETURN $x$.age AS $y$\end{tabular}                                  \\ \hline
\textbf{4) variable relationship}           & $(?x \ \texttt{:knows} \ ?y)$                                                                                        & MATCH ($x$)-{[}:knows{]}-\textgreater{}($y$)                                                                 \\ \hline
\end{tabular}%
}
\end{table}

\paragraph{Pattern type mapping:} Any variable with a class label (pattern \textit{type 1}) can be mapped as labelled node objects in Cypher ($\mathcal{G}_{\mathcal{N}}[s][L] \leftarrow o$). If object is literal (\textit{type 2}), it can be directly mapped as node property value ($\mathcal{G}_{\mathcal{N}}[s][p] \leftarrow o$). Subsequently, whether given triple should map as a node object property (\textit{type 3}) or a relationship between nodes (\textit{type 4}) requires the prior knowledge of whether $?x$ and $?y$ are nodes in the property graph. Thus, Visitor passes WHERE block two times.

\paragraph{1.A) Find nodes} detects all \textit{type 1} triple patterns and classifies their subjects as labelled entity nodes (see alg. \ref{alg:where_pass1}).
\begin{algorithm}[H]
\caption{First pass (find labelled nodes).}
\label{alg:where_pass1}
\begin{algorithmic}
\Require $\mathcal{T}$ \Comment{all triples in WHERE block}
\For{(s, p, o) $\in$ $\mathcal{T}$}
    \If{p $\in$ \{"rdf:type", "a"\}} 
        \State $\mathcal{G}_\mathcal{N}[s][label] \gets o$ 
    \EndIf
\EndFor
\end{algorithmic}
\end{algorithm}

\paragraph{1.B) Categorise patterns} divides all remaining triples ($\mathcal{T}'$) into \textit{type 2}, \textit{type 3} and \textit{type 4}, adding them to $\mathcal{G}_\mathcal{N}$, $\mathcal{G}_{\mathcal{P}}$ and $\mathcal{G}_{\mathcal{R}}$ respectively (see alg. \ref{alg:where_pass2}). 
\begin{algorithm}[H]
\caption{Second pass: Categorise props \& rels.}
\label{alg:where_pass2}
\begin{algorithmic}
\Require $\mathcal{T}'$ (triples), $\mathcal{G}_\mathcal{N}$ (nodes), $\mathcal{G}_\mathcal{V}$ (all vars)
\State $i \gets 0$
\For{(s, p, o) $\in$ $\mathcal{T}'$}
    \If{$s \in \mathcal{G}_\mathcal{N}$ AND $o \in \mathcal{G}_\mathcal{N}$} 
        \State $\mathcal{G}_\mathcal{R}[i] \gets (s, p, o)$ \Comment{add relationship} 
        \State $i \gets i + 1$
    \ElsIf{$o \not \in \mathcal{G}_\mathcal{V}$}
        \State $\mathcal{G}_\mathcal{N}[s][p] \gets o$ \Comment{add prop. constraint}
    \Else
        \State $\mathcal{G}_\mathcal{P}[o] \gets s.p$  \Comment{add var. node property}
    \EndIf
\EndFor
\end{algorithmic}
\end{algorithm}

\subsubsection{Constraints}
\label{subsubsec:constraints}
In SPARQL, constraints ($\mathcal{C}$) are applied on matched patterns using either FILTER or HAVING (Lst. \ref{lst:constraint}).
\begin{lstlisting}[caption=constraint examples,label=lst:constraint]
ex.1: FILTER ?x IN (10,20,30)
ex.2: HAVING (AVG(?y) < 10)
\end{lstlisting}

\paragraph{Categorisation:} While FILTER applies $\mathcal{C}$ before grouping, HAVING applies constraints on grouped results. In contrast, Cypher only has WHERE clause, which can fill the role of both FILTER and HAVING, depending on its position in the query. As discussed in \S\ref{sec:sparql-vs-cypher}, using the WITH clause, we can achieve aggregation and grouping equivalent to SPARQL. See alg. \ref{alg:constraints} for details.

\begin{algorithm}[H]
\caption{Categorise Constraints}
\label{alg:constraints}
\begin{algorithmic}
\Require $\mathcal{C}$ \Comment{constraint in FILTER or HAVING}
\State $e \gets \mathcal{C}.expr$
\State $v \gets \mathcal{C}.var$
\If{\Call{inFILTER}{$\mathcal{C}$}}
    \State $\mathcal{G}_{Where} \gets$ $\mathcal{C}$
\ElsIf{\Call{inHAVING}{$\mathcal{C}$}}
    \If{$v \not\in \mathcal{G}_{WITH}$}
        \State $\mathcal{G}_{WITH}[v] \gets$ "$e$ AS $v$"
    \EndIf
    \State $\mathcal{G}_{WhereWith} \gets$ $\mathcal{C}$ 
\EndIf
\end{algorithmic}
\end{algorithm}

\paragraph{2.A) FILTER:}  If WHERE is applied before any WITH statement, it is equivalent to using FILTER. When $\mathcal{C}$ is in FILTER clause, the full constraint is passed into $\mathcal{G}_{Where}$ as in eq. (\ref{eq:where}). 

\begin{equation}
\label{eq:where}
\begin{split}
    G_{Where} \underset{append}{\leftarrow} [x, \ \texttt{IN}, \ (10, 20, 30)] \\
\end{split}
\end{equation}

\paragraph{2.B) HAVING:}  If WHERE is applied after a WITH statement, it has an equivalent effect to using HAVING.  If $\mathcal{C}(e)$ is within a HAVING clause, the expression ($e$) is aliased in $\mathcal{G}_{WITH} \leftarrow e \ \texttt{AS} \ \texttt{a}$ and the $\mathcal{C}(a)$ is passed to $\mathcal{G}_{WhereWith} \leftarrow \mathcal{C}(a)$ as in eq. (\ref{eq:where_with}).

\begin{equation}
\label{eq:where_with}
\begin{split}
    G_{WITH}[\texttt{agg\_\_0}] & \leftarrow \texttt{AVG}(?y) \ \texttt{AS} \ \texttt{agg\_\_0} \\
    G_{WhereWith} & \underset{append}{\leftarrow} [\texttt{agg\_\_0}, <, 10]
\end{split}
\end{equation}
where alias \lstinline{a=agg__0} is assigned by alg. in \S\ref{alg:visit_aggregate}.

\subsubsection{Projection}
For each expression $e$ in SELECT clause, an equivalent is passed to $\mathcal{G}_{RETURN}$. See \S\ref{alg:visit_select_clause} for details.

\subsubsection{Solution Modifiers}

\paragraph{4.A) GROUP BY} does not have a direct equivalent in Cypher. Instead of explicit GROUP BY, Cypher groups implicitly based on grouping keys in the RETURN or WITH statements. The WITH statement can be used for intermediate grouping or aggregation of results, allowing to further filter the grouped results. See \S\ref{alg:group_condition} for detailed algorithm.

\paragraph{4.B) ORDER BY:} Given a set order conditions  $\mathcal{O}(e)\in \texttt{ORDER BY}$, where $\mathcal{O}$ can be \textit{ASC}, \textit{DESC} or empty and $e$ is either variable or constraint ($\mathcal{C}$), each is added to $\mathcal{G}_{OB} \leftarrow \{e:\mathcal{O}\}$. If the same constraint was used before and is aliased within $\mathcal{G}_{WITH}$ as $a$, then $\mathcal{G}_{OB} \leftarrow a$. See details in \S\ref{alg:order_condition}.

\paragraph{4.C) pagination:} If OFFSET $n$ exists in the SPARQL query, it is added to $\mathcal{G}$ as $\mathcal{G}_{SKIP} \leftarrow n$. If LIMIT $n$ exists in the SPARQL query, it is added to $\mathcal{G}$ as $\mathcal{G}_{LIMIT} \leftarrow n$.

\subsubsection{Event handlers}
\paragraph{VisitAggregate:} Whenever an aggregate function, such as (\Call{COUNT}{?v} AS \textit{?total}) or (\Call{AVG}{?v}), is encountered, it is added to the $\mathcal{G}_{WITH}$ and $\mathcal{G}_{AGG}$ maps using either explicitly assigned aliases (such as \textit{?total}) or implicitly generated aliases (such as \textit{agg\_\_0}). See alg. \ref{alg:visit_aggregate}.

\paragraph{VisitVariable:} Whenever a variable ($v$) is encountered by the Visitor, it is cleaned up (stripped of whitespace and any $?$ or $\$$ characters) and added to the $\mathcal{G}_\mathcal{V}$ set of unique variables. If $v$ is part of an aggregate function, the internal variables are automatically replaced by their equivalent namespaced node property. 

\subsection{Interpreter}
\label{subsec:interpreter}
The Interpreter component is responsible for constructing the final Cypher query from the abstract syntax tree (AST, $\mathcal{G}$) generated by the Visitor (\S\ref{subsec:visitor}). This process involves clause-related classes, each designed to build different parts of the Cypher query based on the structured data in $\mathcal{G}$ (see Tab.\ref{tab:interpreter-components}).

\begin{table}[htb]
\centering
\caption{Key classes of the Interpreter, their functions and the parts of $\mathcal{G}$ they use for the clause construction.}
\label{tab:interpreter-components}
\resizebox{\columnwidth}{!}{%
\begin{tabular}{|l|l|l|}
\hline
\textbf{Class} & \textbf{Function} & \textbf{Utilised AST fields} \\ \hline
MATCHBuilder & Constructs MATCH clause (Pattern matching) & $\mathcal{G}_{\mathcal{N}}$, $\mathcal{G}_{\mathcal{R}}$ \\ \hline
RETURNBuilder & Constructs the RETURN clause (Projection) & $\mathcal{G}_{RETURN}$ \\ \hline
SMBuilder & Constructs ORDER BY, LIMIT and SKIP & $\mathcal{G}_{OB}$, $\mathcal{G}_{LIMIT}$, $\mathcal{G}_{SKIP}$ \\ \hline
WHEREBuilder & WHERE clause (non-aggregated constraints) & $\mathcal{G}_{Where}$ \\ \hline
WITHBuilder & WITH clause (aggregating and grouping) & $\mathcal{G}_{WITH}$ \\ \hline
WWBuilder & second WHERE after WITH (constraints on groups) & $\mathcal{G}_{WhereWith}$ \\ \hline
\end{tabular}%
}
\end{table}

\subsubsection{Building the Cypher Query}
The Interpreter utilises various builder classes to translate $\mathcal{G}$ into a Cypher query. Each builder class is initialised with $\mathcal{G}$ and generates its respective part of the Cypher query. The final Cypher query is assembled by concatenating the outputs of the builders in the correct order (see Listing \ref{lst:parse-ast-to-cypher}).

\begin{lstlisting}[caption=Build Cypher query from given AST ($\mathcal{G}$).,label=lst:parse-ast-to-cypher]
def parse_ast_to_cypher(g: dict):
    match = MATCHBuilder(g).get()
    where = WHEREBuilder(g).get()
    wth = WITHBuilder(g).get()
    where_wth = WWBuilder(g).get()
    unwind = UNWINDBuilder(g).get()
    ret = RETURNBuilder(g).get()
    sm = SMBuilder(g).get_all()

    return "\n".join(c 
        for c in [match, where, 
        wth, where_wth, unwind, 
        ret, sm] if c).strip("\n")
\end{lstlisting}

\paragraph{Interpreter implementation} of the individual Builder classes is detailed in appendix \S\ref{sec:interpreter-implementation}.

\section{Experiments}
\label{sec:experiments}
In the following sections we compare the SPARQL-to-Cypher parsing capabilities of S2CTrans \cite{zhao2023s2ctrans} and our S2CLite parser.
We focus on two types of experiments:

\paragraph{Parsing error rate ($\mathcal{E}$):} We run the parsers on all available SPARQL queries of given dataset and evaluate errors, which result in failure to parse the query into Cypher query (i.e. if no Cypher query was produced, it is accounted as parsing error). See Tab. \ref{tab:parsing-err-legend} for parsing error groups (based on prominent query types) and Tab. \ref{tab:parsing-err-counts} for parsing error results.

\paragraph{Execution accuracy ($\alpha$):} We collect all SPARQL queries that were successfully parsed into Cypher and evaluate mismatch between the outputs when the pair is executed on their respective knowledge graph (SPARQL on RDFlib\footnote{\url{https://rdflib.readthedocs.io/en/stable/}}, Cypher on Neo4j). See Tab. \ref{tab:execution-err-legend} for match error types and Tab. \ref{tab:exec_acc_all} for execution accuracy results. 

\subsection{BSBM42}
\label{subsec:experiments-bsbm42}
As S2CTrans and S2CLite were designed with BSBM42 in mind, both parsed all (42) BSBM42 queries into a syntactically valid Cypher commands (see Tab. \ref{tab:parsing-err-counts}). Furthermore, regarding execution accuracy (Tab. \ref{tab:exec_acc_all}), i.e. the percentage of exact match between results of SPARQL query and parsed Cypher query when executed on their respective knowledge graphs, S2CTrans has no execution errors on BSBM42, while S2CLite has one $\texttt{VAL}$ error, where row values do not match the row in SPARQL result rows. The respective query, where this error occurs is \textit{OptionalRelationship}. We provide the parsed queries, their outputs and analysis in \S\ref{subsubsec:err-bsbm-opt-rel}.

\subsection{Spider4SPARQL}
\label{subsec:experiments-s4s}
Spider4SPARQL dataset contains 6144 usable queries with a significantly larger domain of query types compared to BSBM42. As such, we expect both parsing error rate increase and execution accuracy drop for both S2CTrans and S2CLite. 

\paragraph{Symbols:} Consider $N^{Dev}$ (1032) and $N^{Train}$ (6899) to be the total amount of queries in the respective Spider4SPARQL (S4S) split. $\mathcal{E}^{Split}_{Tool}$ (e.g. $\mathcal{E}^{Train}_{Lite}$) mark the percentage rate of parsing errors (\S\ref{subsubsec:exp-parsing-errors}), while $\alpha^{Split}_{Tool}$ (e.g. $\alpha^{Dev}_{Trans}$) symbolise the execution accuracy (\S\ref{subsubsec:exp-exec-acc}) of the respective S4S splits and parsing tools. Using $\Delta$, change always goes from S2CTrans to S2CLite ($\Delta X=X_{Trans}-X_{Lite}$).

\subsubsection{Parsing errors}
\label{subsubsec:exp-parsing-errors}
Refer to Tab. \ref{tab:parsing-err-counts} for parsing error rate and the counts of parsing errors for individual query group types. We can observe that S2CLite parses significantly more SPARQL queries without any parsing errors, reducing the error rate by $\Delta\mathcal{E}^{Dev}=39.5\%$ and $\Delta\mathcal{E}^{Train}=27.4\%$. S2CTrans is largely limited by not supporting the \lstinline{COUNT(*)} projection aggregation. Both S2CTrans and S2CLite show significant parsing error counts when encountering structured queries like nested SELECT clauses (NS2 in Tab. \ref{tab:parsing-err-counts}) with S2CLite being able to parse about 50\% more into Cypher. Finally, both parsers have comparable error counts when parsing queries with constraint expressions such as \lstinline{NOT EXISTS} and \lstinline{IN} (NS1 in Tab. \ref{tab:parsing-err-counts}), indicating the lack of their support.

\begin{table}[htb]
\centering
\caption{Parsing error types representing query groups.}
\label{tab:parsing-err-legend}
\resizebox{\columnwidth}{!}{%
\begin{tabular}{|r|l|}
\hline
\multicolumn{1}{|c|}{\textbf{error type}} & \textbf{query types included}                                           \\ \hline
COUNT\_ALL                          & queries with COUNT(*) aggregator in SELECT clause                       \\
NS2                                 & query structures (nested SELECT, MINUS, COUNT(*) outside of projection) \\
NS1                                 & queries with constraint expressions (NOT EXISTS, IN/NOT IN, Contains)   \\
OTHER                               & any other unparsed queries                                              \\ \hline
\end{tabular}%
}
\end{table}

\begin{table}[htb]
\centering
\caption{Parsing error types and their counts for BSBM42 and Spider4SPARQL DEV ($N^{Dev}=1032$ queries) and TRAIN sets ($N^{Train}=6899$ queries). See Tab. \ref{tab:parsing-err-legend} for query types included in error categories.}
\label{tab:parsing-err-counts}
\resizebox{\columnwidth}{!}{%
\begin{tabular}{|r|cc|cc|cc|}
\hline
\multicolumn{1}{|l|}{}               & \multicolumn{2}{c|}{\textbf{BSBM (42)}} & \multicolumn{2}{c|}{\textbf{S4S DEV (1032)}} & \multicolumn{2}{c|}{\textbf{S4S TRAIN (6899)}} \\ \cline{2-7} 
\multicolumn{1}{|l|}{\textbf{}}      & \textbf{Trans}      & \textbf{Lite}     & \textbf{Trans}        & \textbf{Lite}        & \textbf{Trans}         & \textbf{Lite}         \\ \hline
\textbf{COUNT\_ALL}                  & 0                   & 0                 & 284                   & \textbf{0}           & 1502                   & \textbf{0}            \\ \hline
\textbf{NS2}                         & 0                   & 0                 & 183                   & \textbf{79}          & 1055                   & \textbf{499}          \\ \hline
\textbf{NS1}                         & 0                   & 0                 & 64                    & \textbf{41}          & 427                    & \textbf{314}          \\ \hline
\textbf{OTHER}                       & 0                   & 0                 & 2                     & 0                    & \textbf{43}            & 63                    \\ \hline
\textbf{err rate $\mathcal{E}$ (\%)} & 0.0\%               & 0.0\%             & 51.6\%                & 11.6\%               & 43.9\%                 & 12.7\%                \\ \hline
\end{tabular}%
}
\end{table}

\begin{figure}[htb]
    \centering
    \includegraphics[width=\columnwidth]{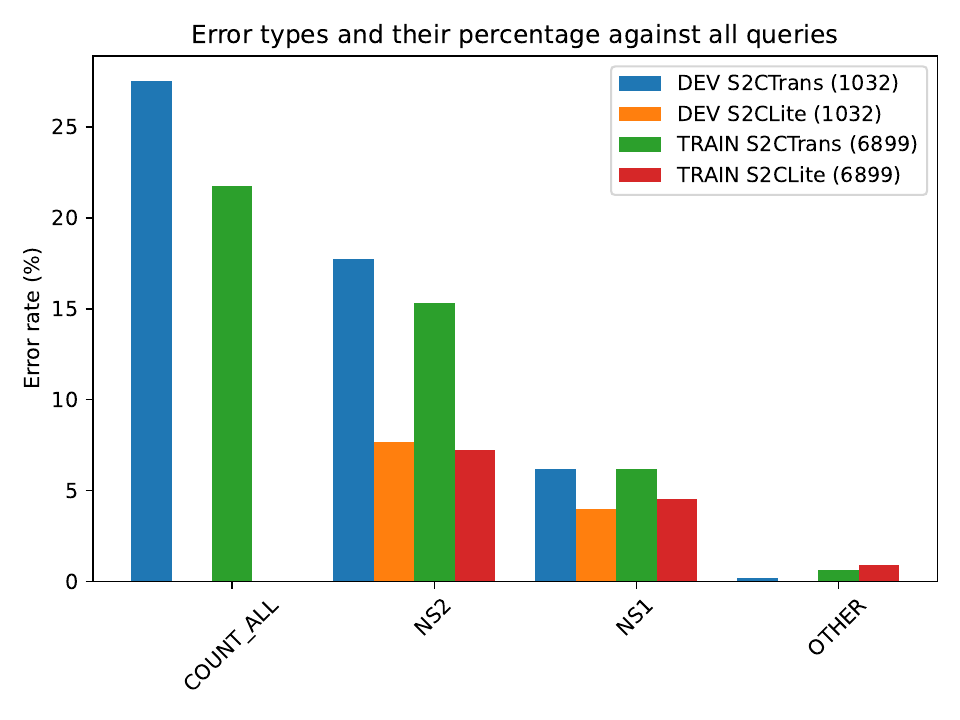}
    \caption{\% of parse errors  against the total number of SPARQL queries passed to the parser ($N^{Dev}=1032$, $N^{Train}=6899$). See Tab. \ref{tab:parsing-err-legend} for error categories.}
    \label{fig:parsing-err-percent}
\end{figure}

\subsubsection{Execution accuracy}
\label{subsubsec:exp-exec-acc}
We collect all the successfully parsed Cypher queries (resp. to the parser tool) and execute them over the Neo4j property graph. The resulting outputs are compared to the bindings produced by the original SPARQL query. We consider the SPARQL and Cypher query outputs to match (to be equivalent) if there are no mismatch errors. We define three types of mismatch errors (see Tab. \ref{tab:execution-err-legend}).

\begin{table}[htb]
\centering
\caption{Types of mismatch errors when comparing SPARQL and Cypher query outputs. When at least one mismatch error is raised, the outputs do not match.}
\label{tab:execution-err-legend}
\resizebox{\columnwidth}{!}{%
\begin{tabular}{|r|l|}
\hline
\multicolumn{1}{|c|}{\textbf{error type}} & \textbf{mismatch error type description}                     \\ \hline
\textbf{NUM\_RES}                         & Number of result rows does not match with SPARQL results     \\
\textbf{VAL}                              & Value of one or more rows does not match with SPARQL results \\
\textbf{N4j\_EXEC}                        & Neo4j raises an error when trying to execute the query       \\ \hline
\end{tabular}%
}
\end{table}

\paragraph{Parsed queries ($C_{\forall}$):} Following experiments in \S\ref{subsubsec:exp-parsing-errors}, we collect all successfully parsed Cypher queries from their respective tools. Out of the $1032$ and $6899$ queries, S2CTrans parsed a total of $499$ and $3545$ queries into Cypher, while S2CLite parsed $980$ and $6487$ respectively (see Tab. \ref{tab:parsed-queries-overview}).

\paragraph{Intersection ($C_{\cap}$):} We also consider a subset of queries, which were parsed successfully by both S2CTrans and S2CLite. This subset accounts to $C^{Dev}_{\cap}=467$ and $C^{Train}_{\cap}=2563$. When compared to $C_{\forall,Trans}$, we can see that most queries parsed by S2CTrans are included in the intersection ($\Delta C^{Dev}_{\forall-\cap}=4$ and $\Delta C^{Train}_{\forall-\cap}=0$). Thus, all but 4 queries parsed by S2CTrans, were also parsed by S2CLite without any parsing errors. 

\begin{table}[htb]
\centering
\caption{Total number of queries ($N$) in respective splits of Spider4SPARQL, total number of Cypher queries parsed without any parsing errors ($C_{\forall}$) and intersection of queries, parsed successfully by both parsers ($C_{\cap}$).}
\label{tab:parsed-queries-overview}
\resizebox{.55\columnwidth}{!}{%
\begin{tabular}{|r|cc|cc|}
\hline
\multicolumn{1}{|l|}{} & \multicolumn{2}{c|}{\textbf{S4S DEV}} & \multicolumn{2}{c|}{\textbf{S4S TRAIN}} \\ \cline{2-5} 
\multicolumn{1}{|l|}{} & \textbf{Trans}     & \textbf{Lite}    & \textbf{Trans}      & \textbf{Lite}     \\ \hline
\textbf{$N$}           & \multicolumn{2}{c|}{1032}             & \multicolumn{2}{c|}{6899}               \\ \hline
\textbf{$C_{\forall}$} & 499                & 912              & 3545                & 5790              \\ \hline
\textbf{$C_{\cap}$}    & \multicolumn{2}{c|}{495}              & \multicolumn{2}{c|}{3541}               \\ \hline
\end{tabular}%
}
\end{table}

\begin{figure}[htb]
    \centering
    \includegraphics[width=.95\columnwidth]{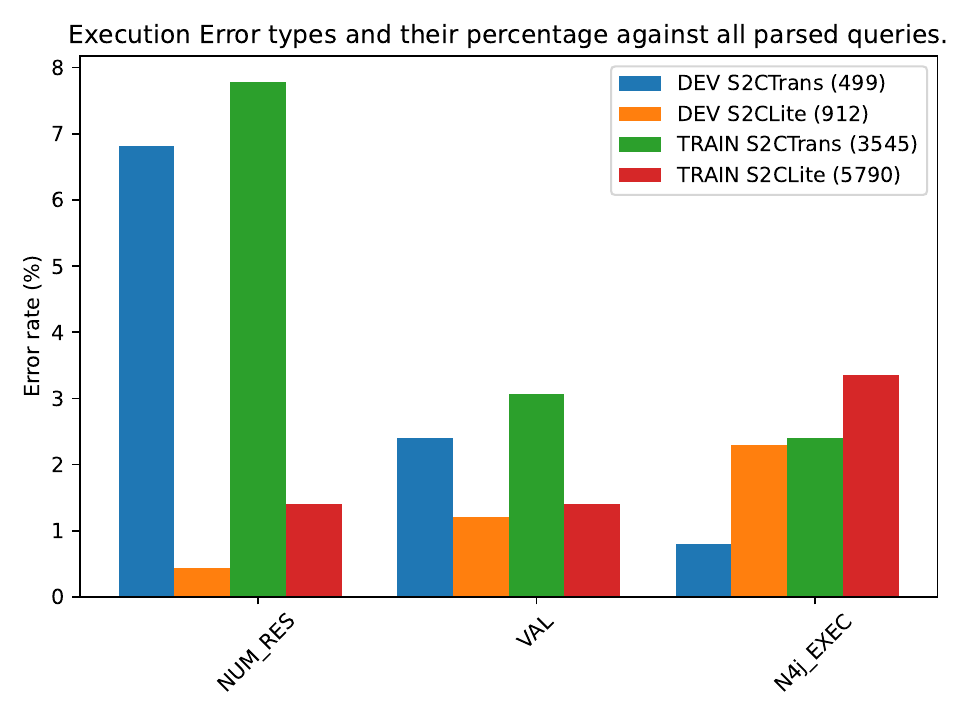}
    \caption{Execution error rate (\%) against the total number of \textbf{all} successfully parsed Cypher queries ($C_{\forall}$).}
    \label{fig:exec-err-percent}
\end{figure}

\begin{figure}[htb]
    \centering
    \includegraphics[width=.95\columnwidth]{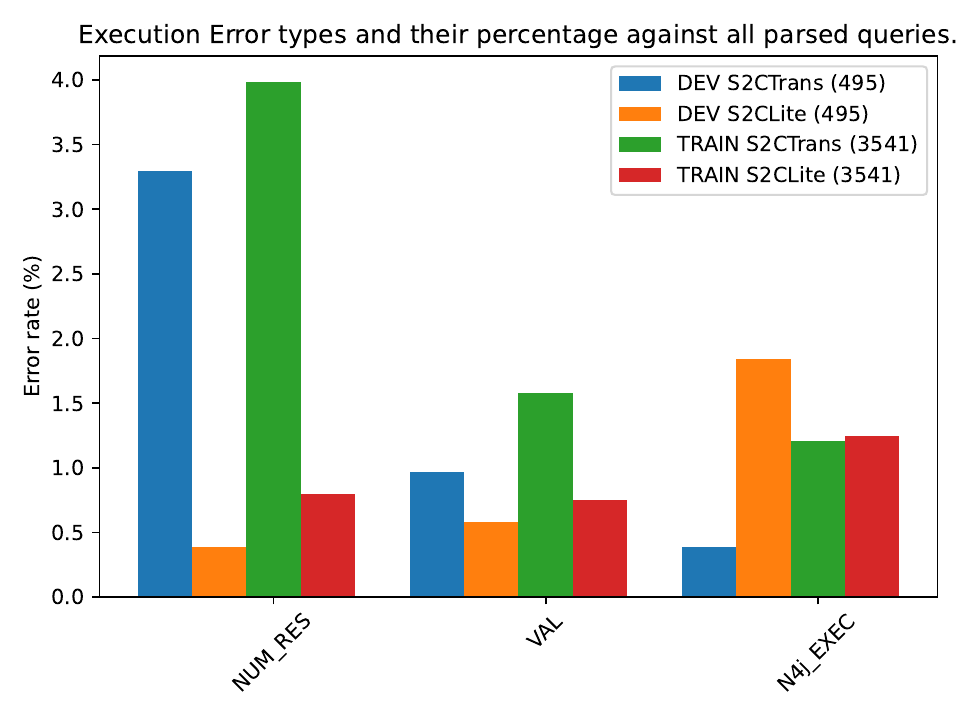}
    \caption{Execution error rate (\%) against the total count of \textbf{intersecting} parsed Cypher queries ($C_{\cap}$).}
    \label{fig:S4S_intersection_exec-err-percent}
\end{figure}

\paragraph{Execution accuracy:} For a fair comparison of execution accuracy ($\alpha$) between S2CTrans and S2CLite, we conduct two experiments:
\begin{enumerate}
    \item $\alpha_{\forall}$: Execute all available parsed Cypher queries ($C_{\forall}$), count their match errors and calculate relative $\alpha$ against $C_{\forall}$ and total accuracy ($\tau$) against $N$  (Tab. \ref{tab:exec_acc_all} and Fig. \ref{fig:exec-err-percent}).
    \item $\alpha_{\cap}$: Repeat the above for only intersecting queries ($C_{\cap}$), parsed successfully by both S2CTrans and S2CLite (Tab. \ref{tab:S4S_intersection_exec_err} and Fig. \ref{fig:S4S_intersection_exec-err-percent}).
\end{enumerate}

\paragraph{Discussion:} Given the execution match ($M$) in Tab. \ref{tab:exec_acc_all}, S2CLite in its current state successfully transforms 4779 out of the total of 6144 given SPARQL queries ($N^{Dev}+N^{Train}$), amounting to \textbf{total parsing accuracy} $\tau_{Lite}=77.8$\%  (compared to $\tau_{Trans}=44.2\%$) on Spider4SPARQL dataset.

\paragraph{Considering $\alpha_{\forall}$} in Tab. \ref{tab:exec_acc_all}, we observe higher execution accuracy in S2CTrans for both DEV and TRAIN, with significant increase in TRAIN ($\Delta \alpha^{Train}_{\forall}=6.1\%$). This suggests robust parsing rules and well defined boundaries of S2CTrans, limiting itself only to queries it was designed for. In parallel, Fig. \ref{fig:exec-err-percent} shows that only a minimal portion of match errors in S2CTrans stem from query execution errors (N4j\_EXEC), while ca. 7\% of parsed queries, return different number of results (NUM\_RES) compared to the SPARQL bindings. Conversely, for S2CLite 8 to 13\% (DEV, TRAIN resp.) of its parsed Cypher queries fail to execute properly. This suggests that S2CLite generates false positives (syntactically invalid Cypher queries) in the parsing stage.     

\paragraph{Considering $\alpha_{\cap}$} in Tab. \ref{tab:S4S_intersection_exec_err}, S2CTrans shows lower execution accuracy in both dataset splits ($\Delta \alpha^{Dev}_{\cap}=-6.0\%$ and $\Delta \alpha^{Train}_{\cap}=-7.3\%$). Subsequently, noting a significant reduction of match error rates for S2CLite in Fig. \ref{fig:S4S_intersection_exec-err-percent}, we conclude that S2CLite produces more semantically accurate queries than S2CTrans, when restricted to $C_{\cap}$. 

\begin{table}[htb]
\centering
\caption{Execution accuracy ($\alpha_{\forall}$) of \textbf{all} parsed queries. $C_{\forall}$ marks the count of all parsed Cypher queries. $M$ counts how many $C_{\forall}$ had their result projections match with the SPARQL query results. \textbf{err} rows show counts of execution errors (see Tab. \ref{tab:execution-err-legend}). Final rows list execution accuracy ($\alpha_{\forall}=\frac{M}{C_{\forall}}$) and total accuracy ($\tau=\frac{M}{N}$).}
\label{tab:exec_acc_all}
\resizebox{\columnwidth}{!}{%
\begin{tabular}{|r|cc|cc|cc|}
\hline
\multicolumn{1}{|l|}{}                    & \multicolumn{2}{c|}{\textbf{BSBM (42)}} & \multicolumn{2}{c|}{\textbf{S4S DEV (1032)}} & \multicolumn{2}{c|}{\textbf{S4S TRAIN (6899)}} \\ \cline{2-7} 
\multicolumn{1}{|l|}{\textbf{}}           & \textbf{Trans}      & \textbf{Lite}     & \textbf{Trans}       & \textbf{Lite}         & \textbf{Trans}        & \textbf{Lite}          \\ \hline
\textbf{parsed ($C_{\forall}$)}           & 42                  & 42                & 499                  & \textbf{912}          & 3545                  & \textbf{5790}          \\
\textbf{exec match ($M$)}                 & \textbf{42}         & 41                & 449                  & 876                   & 3075                  & 5434                   \\ \hline
\textbf{err (NUM\_RES)}                   & 0                   & 0                 & 34                   & \textbf{4}            & 276                   & \textbf{81}            \\
\textbf{err (VAL)}                        & \textbf{0}          & 1                 & 12                   & \textbf{11}           & 109                   & \textbf{81}            \\
\textbf{err (N4j\_EXEC)}                  & 0                   & 0                 & \textbf{4}           & 21                    & \textbf{85}           & 194                    \\ \hline
\textbf{exec acc $\alpha_{\forall}$ (\%)} & \textbf{100\%}      & 97.6\%            & 90.0\%               & \textbf{96.1\%}       & 86.7\%                & \textbf{93.9\%}        \\
\textbf{total acc $\tau$ (\%)}            & \textbf{100\%}      & 97.6\%            & 43.5\%               & \textbf{84.9\%}       & 44.6\%                & \textbf{78.8\%}        \\ \hline
\end{tabular}%
}
\end{table}

\begin{table}[htb]
\centering
\caption{Execution accuracy ($\alpha_{\cap}$) of \textbf{intersecting} parsed queries. $C_{\cap}$ marks queries successfully parsed by both parsers. $M$ counts how many of $C_{\cap}$ results match with the SPARQL query results. Further rows count error types (see Tab. \ref{tab:execution-err-legend}). Final row shows $\alpha_{\cap}=\frac{M}{C_{\cap}}$. }
\label{tab:S4S_intersection_exec_err}
\resizebox{.8\columnwidth}{!}{%
\begin{tabular}{|r|cc|cc|}
\hline
\multicolumn{1}{|l|}{}                 & \multicolumn{2}{c|}{\textbf{S4S DEV (1032)}} & \multicolumn{2}{c|}{\textbf{S4S TRAIN (6899)}} \\ \cline{2-5} 
\multicolumn{1}{|l|}{\textbf{}}        & \textbf{Trans}       & \textbf{Lite}         & \textbf{Trans}        & \textbf{Lite}          \\ \hline
\textbf{parsed ($C_{\cap}$)}           & 495                  & 495                   & 3541                  & 3541                   \\
\textbf{exec match ($M$)}              & 447                  & 466                   & 3074                  & 3348                   \\ \hline
\textbf{err (NUM\_RES)}                & 34                   & \textbf{4}            & 275                   & \textbf{55}            \\
\textbf{err (VAL)}                     & 10                   & \textbf{6}            & 109                   & \textbf{52}            \\
\textbf{err (N4j\_EXEC)}               & \textbf{4}           & 19                    & \textbf{83}           & 86                     \\ \hline
\textbf{exec acc $\alpha_{\cap}$ (\%)} & 90.3\%               & \textbf{94.1\%}       & 86.8\%                & \textbf{94.5\%}        \\ \hline
\end{tabular}%
}
\end{table}

\section{Conclusion}
\label{sec:conclusion}
We presented S2CLite (\S\ref{sec:methods}), a stand-alone algorithmic SPARQL to Cypher (S2C) parser, as a lightweight alternative to existing S2C parsers (\S\ref{subsec:s2c-parsers}) and evaluated its feasibility on BSBM42 (\S\ref{subsubsec:ds-bsbm42}) and Spider4SPARQL (S4S, \S\ref{subsubsec:ds-s4s}) datasets. In our experiments (\S\ref{sec:experiments}), we compare S2CLite to the current SotA parser S2CTrans (\S\ref{subsubsec:s2ctrans}). We conclude that S2CLite outperforms S2CTrans both in terms of parsing errors (\S\ref{subsubsec:exp-parsing-errors}) and execution accuracy (\S\ref{subsubsec:exp-exec-acc}). In \S\ref{subsubsec:exp-parsing-errors} we show that S2CLite parses a larger subset of query types (ca. 33\% error decrease, Tab. \ref{tab:parsing-err-counts}) but produces significantly more Cypher queries with execution errors (N4j\_EXEC in Tab. \ref{tab:exec_acc_all}). However, when limited to the intersection of queries between S2CTrans and S2CLite ($C_{\cap}$, Tab. \ref{tab:parsed-queries-overview}), S2CLite offers 6\% execution accuracy increase over S2CTrans (Tab. \ref{tab:S4S_intersection_exec_err}). As a use case, we utilized S2CLite to efficiently parse Spider4SPARQL queries to Cypher and generate the joint Spider4SSC dataset (\S\ref{app:s4ssc}). 



\newpage

\section{Limitations}
\label{sec:limitations}
\paragraph{Explicit relationships:} As we discuss in \S\ref{app:explcit-rel-type-names}, while S2CLite can directly translate any given SPARQL query to Cypher without any third-party dependencies, it is limited in the ability of differentiating between property and relationship patterns between variable nodes. This limitation is alleviated by allowing to explicitly define a set of predicates, which are to be handled as node relationships (shown as $R_e$ in Fig. \ref{fig:s2clite-pipeline}). However, an alternative approach may be desirable, if a file or ontology of the knowledge graph is available at hand.

\paragraph{Nested queries:} During the parsing error rate evaluation of S2CLite, we show that ca. 8\% of the Spider4SPARQL dataset queries are not supported due to the existence of nested queries (see NS2 in Fig. \ref{fig:parsing-err-percent}). Our aim is to implement support for simple nested queries to eliminate this limitation and further increase the parsing accuracy.

\paragraph{False positives:} Comparing the error counts between Tab. \ref{tab:exec_acc_all} and Tab. \ref{tab:S4S_intersection_exec_err} indicates a drastic reduction in S2CLite N4j\_EXEC errors, suggesting that many of the false positives (queries that are parsed but not executable) are the ones that are outside of the $C_{\forall}$ intersection subset. This calls for better definition of explicit constraints and boundaries on the queries that S2CLite supports. In future versions of S2CLite, we aim to include concrete error messages, when unsupported queries are encountered. 

\paragraph{MATCH after WHERE:} As we show in \ref{subsec:experiments-bsbm42}, S2CLite fails to produce exact match of results for the \textit{OptionalRelationship} SPARQL query (see \S\ref{subsubsec:err-bsbm-opt-rel} for details). This may stem from a known limitation of S2CLite and needs to be addressed further. S2CLite always positions all MATCH patterns at the beginning of the query and does not define logic for positioning MATCH patterns after WHERE, which is required, in this scenario, to produce the exact results match to the SPARQL query execution.
mapped 
\paragraph{Limited dataset testing:} One of the foundational aims of S2CLite is the ability to, given a large scale Text-to-SPARQL dataset, produce structurally and semantically identical Text-to-Cypher datasets. By the nature of Text-to-QL, the entries in such datasets need to be covering many domains and query types, which asks for further tests on different Text-to-SPARQL datasets.

\paragraph{Unique queries:} Our experiments consider all valid entries in the Spider4SPARQL dataset as queries. However, similar to Spider, Spider4SPARQL may contain multiple entries with the same query, paraphrasing the question. Thus, the query counts in our experiments include these duplicate queries.

\paragraph{Reduced Spider4SSC dataset size:} The generated Spider4SSC dataset is currently limited to 4525 unique entries (4496 unique questions out of the total 7893). This is due to our strict selection process, where we only include (question, sql, sparql, cypher) entries, which return exactly the same results when executed on their respective database. See \ref{tab:s4ssc-unq-questions} for our count of questions with exactly matching (\textit{equivalent}) query execution results. The \textit{not equivalent} entries have diverse causes, such as no result from one of the languages, same fields but different values, different number of entries,  extra fields bound at the output etc. In the future, we plan to use general LLMs to attempt at fixing mistakes in the queries to further increase the size of the \textit{equivalent} subset and expanding Spider4SSC with newly acquired matching queries. 

\begin{table}[htb]
\centering
\caption{Counts of unique questions with equivalent, not equivalent or empty execution results across all three query languages when executed on their respective database. We use the only the **equivalent** questions to build our Spider4SSC dataset.}
\label{tab:s4ssc-unq-questions}
\resizebox{.8\columnwidth}{!}{%
\begin{tabular}{|l|ll|l|}
\hline
\textbf{unique questions} & \textbf{dev} & \textbf{train} & \textbf{total} \\ \hline
\textbf{equivalent}       & 608          & 3888           & 4496           \\
\textbf{not equivalent}   & 402          & 2817           & 3219           \\
\textbf{empty}            & 22           & 156            & 178            \\ \hline
\textbf{total}            & 1032         & 6861           & 7893           \\ \hline
\end{tabular}%
}
\end{table}

\section{Potential risks}
S2CLite is a rule-based parser with limited query support, which was generally aimed to accommodate the available query types in BSBM, Spider and Spider4SPARQL datasets. It is not guaranteed that any arbitrary SPARQL query will be parsed into semantically equivalent Cypher query. With this in mind, if S2CLite is used for automatic dataset creation, further evaluation of the query equivalence such as execution accuracy should be conducted.

\newpage

\bibliography{anthology,custom}

\appendix
\clearpage
\newpage

\section{Spider4SSC}
\label{app:s4ssc}
Spider4SSC (download\footnote{\url{https://www.dropbox.com/scl/fi/v4wa82hpe2s5p5uh4bgnj/Spider4SSC.zip?rlkey=rb447tkhjq38ugnd0v8rpt2bh&e=3&st=s94sdngb}}) is a composite Text-to-Query language dataset with 4525 unique (question, sql, sparql, cypher) entries where semantically matching queries are provided in SQL, SPARQL and Cypher. To build the dataset, we downloaded the Spider4SPARQL dataset dev and train JSONs and materialised RDF knowledge graphs (KGs) following the instructions on \citeposs{kosten2023spider4sparql} GitHub\footnote{\url{https://github.com/ckosten/Spider4SPARQL}}. Furthermore, we fixed and re-materialised 23 empty KGs and omitted 7 with no SQLite counterpart, resulting in total of 159 valid and equivalent SQLite/RDF graph databases. Using S2CLite (\S\ref{sec:methods}), we parsed all available SPARQL queries in Spider4SPARQL into matching Cypher queries. Finally, downloading Spider queries and SQLite databases\footnote{\url{https://yale-lily.github.io/spider}}, we assembled the \textbf{Spider4SSC} dataset from all executable and valid queries. To ensure the queries are equivalent, all three queries must return matching results when executed over their respective database (using sqlite3\footnote{\url{https://docs.python.org/3.11/library/sqlite3.html}} for SQL, rdflib 7.0.0\footnote{\url{https://rdflib.readthedocs.io/en/stable/}} for SPARQL and Neo4j with neosemantics\footnote{\cite{barrasa2021neosemantics}} for Cypher). We include only question/query pairs which strictly adhere to this rule. Refer to Listing \ref{lst:s4ssc-example} for an example entry from Spider4SSC dev set. A full technical report detailing the process of creating Spider4SSC will be published at a later time on arXiv.

\begin{lstlisting}[basicstyle=\footnotesize, caption={Sample entry from Spider4SSC dataset.}, label=lst:s4ssc-example]
{
"db_id": "concert_singer",
"question": "How many singers do we have?",
"sql": "SELECT count(*) FROM singer",
"sparql": "select (count( *) as ?aggregation_all) 
           where { ?t1 a :singer . }",
"cypher": "MATCH (t1:ROOT__singer) 
           WITH COUNT(*) AS aggregation_all 
           RETURN aggregation_all",
"namespaces": [
    "concert",
    "stadium",
    "singer_in_concert",
    "singer",
    "XMLSchema"
]
},
\end{lstlisting}

\section{SPARQL parse tree (sparkle-g)}
\label{app:sparkle-g}
Parse tree ($T_s$) is an ordered tree that formalises the syntactic structure of a given string using specific grammar \cite{blackburn-etal-1993-talking}. For SPARQL string and grammar, each non-terminal (non-leaf) node in $T_s$ represents SPARQL constructs (clause, pattern or function, while terminal (leaf) nodes are reserved for symbols (keywords, literals, variables, operators, etc.). In our work, we use SPARQL 1.1 grammar from the sparkle-g project \cite{2013sparkleg}, which adheres to the official W3C SPARQL 1.1 grammar\footnote{\url{https://www.w3.org/TR/sparql11-query/\#grammar}}. We use version 4 of the public-domain parser generator ANTLR \cite{parr1995antlr} to compile a Python-based SPARQL 1.1 grammar parser. We further refer to this compiled parser simply as \textbf{sparkle-g}, to honour the original authors. There are several other available options for implementation of SPARQL-based grammar, such as SPARQL 1.0 grammar from the ANTLR4 grammars repository\footnote{\url{https://github.com/antlr/grammars-v4/tree/master}} or SPARQL 1.1 grammar from Martin Schröder\footnote{\url{https://github.com/schrer/sparql-g4}}.

\subsection{SPARQL parse tree example}
\begin{figure*}
    \centering
    \includegraphics[height=0.8\paperheight]{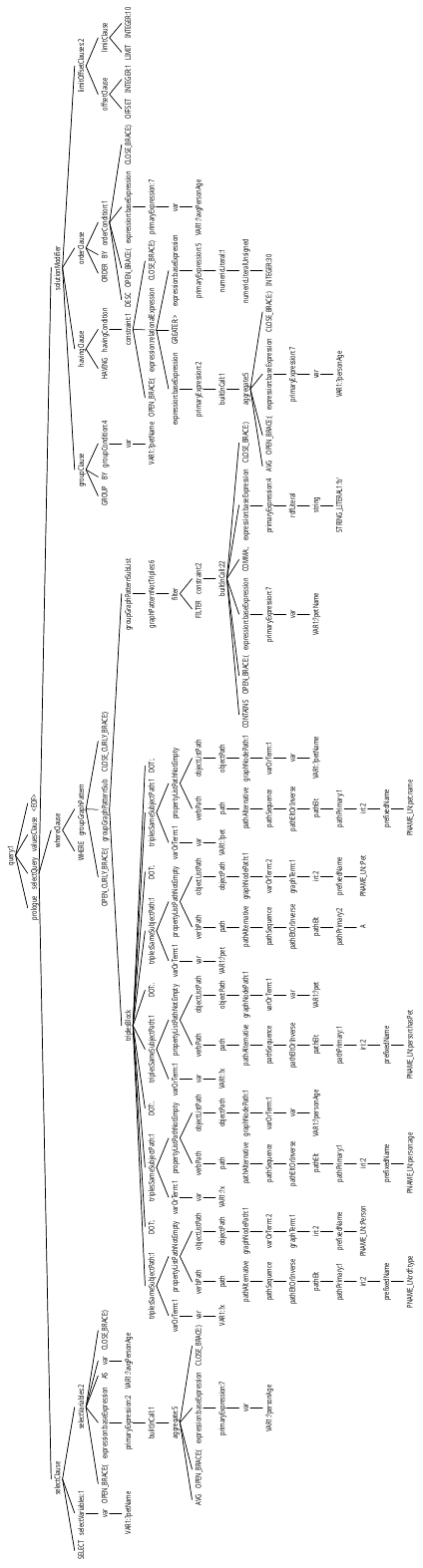}
    \caption{A parse tree of SPARQL clause in Fig. \ref{fig:visitor-overview} (and Lst. \ref{lst:sparql-query-01}), generated using sparkle-g grammar (\S\ref{app:sparkle-g}).}
    \label{fig:sparql-query-01-parse-tree}
\end{figure*}

Listing \ref{lst:sparql-query-01} shows the SPARQL query from Fig. \ref{fig:visitor-overview}. The generated parse tree using ANTLR4 parser with sparkle-g grammar is shown in Fig. \ref{fig:sparql-query-01-parse-tree}. 

\begin{lstlisting}[caption={SPARQL query example (same as in Fig. \ref{fig:visitor-overview}).}, label=lst:sparql-query-01]
SELECT ?petName (AVG(?personAge) 
AS ?avgPersonAge)
WHERE { 
  ?x rdf:type :Person .
  ?x person:age ?personAge .
  ?x person:hasPet ?pet .
  ?pet a :Pet .
  ?pet pet:name ?petName .
  FILTER CONTAINS(?petName, 'b')
}
GROUP BY ?petName
HAVING (AVG(?personAge) > 30)
ORDER BY DESC(?avgPersonAge)
OFFSET 1
LIMIT 10
\end{lstlisting}

\newpage




\section{Explicit relationship type names}
\label{app:explcit-rel-type-names}
In \ref{fig:s2clite-pipeline} we mention that Visitor can receive optional list of explicit relationships ($R_e$). To understand the necessity for this, consider the example query in Listing \ref{lst:ex1}.

\begin{lstlisting}[caption={\textit{Average age of all people who Emma knows.}}, label=lst:ex1]
PREFIX : <http://example.org/>
SELECT AVG(?ag) AS ?avgAge
WHERE { 
  ?x a :Person .
  ?x :name 'Emma' .
  ?x :knows ?y .
  ?y :age ?ag .
}
\end{lstlisting}
Note the omission of class definition (type 1 pattern) for ?y (i.e. \lstinline{(?y a :Person})). This is a valid SPARQL in which \lstinline{(?x :knows ?y)} represents a type 4 pattern (see Tab. \ref{tab:s2c-pattern-matching}). However, as we discuss in \ref{subsubsec:pattern-matching}, alg. \ref{alg:where_pass2} identifies pattern as type 4 only when \lstinline{?x} and \lstinline{?y} are both known node object variables. As alg. \ref{alg:where_pass1} will not mark \lstinline{?y} as node object, the pattern \lstinline{(?x :knows ?y)} will get categorised as type 3 pattern instead.

\paragraph{Explicit relationship types} circumvent the default pattern categorisation. Any predicate specified in $R_e$ will automatically set pattern as type 4. For example, if $R_e=(\texttt{':knows'},)$, then \lstinline{(?x :knows ?y)} is explicitly type 4 and is added to $\mathcal{G}_{\mathcal{R}}$ instead of $\mathcal{G}_{\mathcal{P}}$.  

\newpage

\section{Visitor algorithm details}
Bellow are pseudocode algorithms of the parsing strategy implemented in our custom Visitor class, which traverses the SPARQL parse tree and builds an abstract syntax tree (AST) represented as a graph $\mathcal{G}$ (Python dictionary class in practice). The graph $\mathcal{G}$ is then used to assemble an equivalent Cypher query. Algorithms used in the main sections are also included again for completeness.

\subsection{Initialise AST Containers}
\label{alg:init}
The graph containers are initialised to store various elements encountered during the parsing.

\begin{algorithm}[H]
\caption{Initialise graph containers}
\begin{algorithmic}
\Ensure $\mathcal{G} \gets \{\}$  \Comment{empty graph}
\State $\mathcal{G}_\mathcal{V} \gets [\ ]$ \Comment{empty variable list} 
\State $\mathcal{G}_\mathcal{N} \gets \{\}$ \Comment{empty node map}
\State $\mathcal{G}_\mathcal{P} \gets \{\}$ \Comment{empty node property map}
\State $\mathcal{G}_\mathcal{R} \gets [\ ]$ \Comment{empty relationship list}
\State $\mathcal{G}_{T(\mathcal{R})} \gets [\ ]$ \Comment{empty relationship types list}
\State $\mathcal{G}_{AGG} \gets \{\}$ \Comment{empty map of aggregation functions}
\State $\mathcal{G}_{SUBG} \gets \{\}$ \Comment{empty map of subgraphs (not implemented)}
\State $\mathcal{G}_{Where} \gets [\ ]$ \Comment{empty list of simple constraints}
\State $\mathcal{G}_{With} \gets \{\}$ \Comment{empty map of aliases}
\State $\mathcal{G}_{WhereWith} \gets [\ ]$ \Comment{empty list of aggregated/aliased constraints}
\State $\mathcal{G}_{UNWIND} \gets \{\}$ \Comment{empty map of collections to unwind}
\State $\mathcal{G}_{RETURN} \gets [\ ]$ \Comment{empty list of variables to return}
\State $\mathcal{G}_{OB} \gets \{\}$ \Comment{empty map of variables to order by}
\State $\mathcal{G}_{LIMIT} \gets null$ \Comment{int container for limiting returned rows}
\State $\mathcal{G}_{SKIP} \gets null$ \Comment{int container for skipping returned rows}
\end{algorithmic}
\end{algorithm}

\newpage

\subsection{Visit WHERE Clause}
The WHERE clause is visited in two passes. The first pass identifies which resources are nodes and labels them. The second pass categorises triples into node properties and relationships.

\subsection{First Pass: Identify Entities and Labels}
\label{app:where_pass1}
\begin{algorithm}[H]
\caption{WHERE Clause Pass 1: Identify Labelled Nodes}
\begin{algorithmic}
\Require $\mathcal{T}$ \Comment{all triples in query}
\For{(s, p, o) $\in$ $\mathcal{T}$}
    \If{p $\in$ \{"rdf:type", "a"\}} 
        \State $\mathcal{G}_\mathcal{N}[s][label] \gets o$ 
    \EndIf
\EndFor
\end{algorithmic}
\end{algorithm}

\subsection{Second Pass: Categorise Properties and Relationships}
\label{app:where_pass2}

\begin{algorithm}[H]
\caption{WHERE Clause Pass 2: Categorise Properties and Relationships}
\begin{algorithmic}
\Require $\mathcal{T}'$ ... triples, $\mathcal{G}_\mathcal{N}$ ... nodes, $\mathcal{G}_\mathcal{V}$ ... all variables
\State $i \gets 0$
\For{(s, p, o) $\in$ $\mathcal{T}'$}
    \If{$s \in \mathcal{G}_\mathcal{N}$ AND $o \in \mathcal{G}_\mathcal{N}$} 
        \State $\mathcal{G}_\mathcal{R}[i] \gets (s, p, o)$ \Comment{add relationship} 
        \State $i \gets i + 1$
    \ElsIf{$o \not \in \mathcal{G}_\mathcal{V}$}
        \State $\mathcal{G}_\mathcal{N}[s][p] \gets o$ \Comment{add node property constraint}
    \Else
        \State $\mathcal{G}_\mathcal{P}[o] \gets s.p$  \Comment{add variable node property}
    \EndIf
\EndFor
\end{algorithmic}
\end{algorithm}

\subsection{Categorise Constraints}
\label{app:constraints}
Constraints are divided into two groups. First group accounts for FILTER clauses with no aggregation needed. Second group contains those that operate on projected solutions, amounting to any HAVING clauses or FILTER clauses containing aggregation functions.

\begin{algorithm}[H]
\caption{visitConstraint}
\begin{algorithmic}
\Require $\mathcal{C}$ \Comment{constraint in FILTER or HAVING}
\State $e \gets \mathcal{C}.expr$
\State $v \gets \mathcal{C}.var$
\If{\Call{inFILTER}{$\mathcal{C}$}}
    \State $\mathcal{G}_{Where} \gets$ $\mathcal{C}$
\ElsIf{\Call{inHAVING}{$\mathcal{C}$}}
    \If{$v \not\in \mathcal{G}_{WITH}$}
        \State $\mathcal{G}_{WITH}[v] \gets$ "$e$ AS $v$"
    \EndIf
    \State $\mathcal{G}_{WhereWith} \gets$ $\mathcal{C}$ 
\EndIf
\end{algorithmic}
\end{algorithm}

\subsection{Visit SELECT}
\label{alg:visit_select_clause}
The SELECT clause contains all the variables and aliased aggregations required as the output of the query. As Visitor walks every child node, any handling functions such as visitVar (\S\ref{alg:visit_var}) or visitAggregate (\S\ref{alg:visit_aggregate}) are also triggered.

\begin{algorithm}[H]
\caption{visitSelectClause}
\begin{algorithmic}
\Require $C$ \Comment{context of the SELECT clause}
\State $V \gets [\ ]$ \Comment{list of projection variables}
\State $K \gets [\ ]$ \Comment{list of keywords}
\State $d \gets False$ \Comment{DISTINCT flag}
\For{$ch \in \Call{getChildren}{C}$}
    \If{\Call{isTerminal}{$ch$}}
        \State $t \gets \Call{upper}{ch.text}$
        \If{$t$ == "DISTINCT"}
            \State $d \gets True$
        \ElsIf{$t$ == "*"}
            \State $V \gets$ "*"
        \Else
            \State $K \gets t$   
        \EndIf
    \Else
        \State $t_{ch} \gets \Call{visit}{ch}$
        \If{$d == True$}
            \State $V \gets $"DISTINCT $t_{ch}$"
            \State $d \gets False$
        \Else
            \State $V \gets t_{ch}$
        \EndIf
    \EndIf
\EndFor
\State $V_{str} \gets$ ", ".join($V$) 
\State $\mathcal{G}_{RETURN}$.extend($K$)
\State $\mathcal{G}_{RETURN}$.append($V_{str}$)
\end{algorithmic}
\end{algorithm}

\subsection{Solution Modifiers}
The Solution modifier methods further process the output variables, handling grouping (\S\ref{alg:group_condition}), ordering (\S\ref{alg:order_condition}) and pagination of result rows with LIMIT (\S\ref{alg:visit_limit}) and OFFSET (\S\ref{alg:visit_offset}).

\newpage

\subsubsection{Visit GROUP BY}
\label{alg:group_condition}
Whenever GROUP BY is encountered, there can be one or more order conditions $c_{GB}\in C_{GB}$. Each $c_{GB}$ can be one of
\begin{enumerate}
    \item $c_{GB} = e \texttt{ [AS }?v\texttt{]}$ (expression \& opt. alias)
    \item $c_{GB} = f(e)$ (built in or custom function call)
    \item $c_{GB} = ?v$ (variable)
\end{enumerate}
Alg. \ref{alg:visitGroupCondition} ensures that for each $c_{GB}\in C_{GB}$, group variables are added to the WITH clause and any other RETURN variables, that are not present in $\mathcal{G}_{WITH}$ are called by their name-spaced node property (e.g. $node.name$ instead of $?nodeName$).

\begin{algorithm}[H]
\caption{visitGroupCondition}
\label{alg:visitGroupCondition}
\begin{algorithmic}
\Require $c_{GB}$ \Comment{one condition in GROUP BY}
\State $g \gets [\ ]$ \Comment{group list}
\State $f_{\mathcal{A}} \gets \text{False}$ \Comment{flag for alias presence}

\For{$ch \in \Call{getChildren}{c_{GB}}$}
    \If{$\Call{isTerminal}{ch}$}
        \State $t \gets \Call{upper}{ch.text}$
    \Else
        \State $t \gets \Call{visit}{ch}$
    \EndIf

    \If{$t \in $["(", ")"]}
        \State continue \Comment{ignore any brackets}
    \ElsIf{$t == "AS"$}
        \State $f_{\mathcal{A}} \gets \text{True}$
    \EndIf

    \State $g \underset{app}{\gets} t$ \Comment{append to $g$}
\EndFor

\If{$f_{\mathcal{A}}$ == True}
    \State $\mathcal{A} \gets \Call{pop}{g, -1}$ \Comment{pop last element as alias}
    \State $v \gets g[1]$
    \If{$v \notin \Call{keys}{\mathcal{G}_{WITH}}$}
        \State $\mathcal{A}_{dict} \gets \{\mathcal{A}: v \text{ AS } \mathcal{A}\}$
        \State $\Call{update}{\mathcal{G}_{WITH}, \mathcal{A}_{dict}}$
        \State $g[1] \gets \mathcal{A}$
    \EndIf
    \State \Call{pop}{g, -1} \Comment{remove AS}
    \State $f_{\mathcal{A}} \gets \text{False}$
\EndIf

\State $g_{str} \gets \Call{join}{\text{" "}, g}$
\If{$v \in \Call{keys}{\mathcal{G}_{PROPS}}$}
    \State $\mathcal{G}_{WITH}[v] \gets \mathcal{G}_{PROPS}[v] \text{ AS } v$
\EndIf

\State \Return $g_{str}$
\end{algorithmic}
\end{algorithm}

\newpage

\subsubsection{Visit ORDER BY}
When ORDER BY is present in the query, there can be one or more order conditions $c\in C_{OB}$. Each $c$ can be one of
\begin{enumerate}
    \item $c = \texttt{(ASC|DESC)(}e\texttt{)}$ (order and expression)
    \item $c = \mathcal{C}$ (constraint)
    \item $c = ?v$ (variable)
\end{enumerate}
Alg. \ref{alg:visitOrderCondition} applies for each order condition $c_{OB} \in C_{OB}$ and extends the $\mathcal{G}_{OB}$ map, using either existing alias from $\mathcal{G}_{WITH}[c_{OB}]$ or simply passing $c_{OB}$ as is.
\label{alg:order_condition}
\begin{algorithm}[H]
\caption{visitOrderCondition}
\label{alg:visitOrderCondition}
\begin{algorithmic}
\Require $c_{OB}$ \Comment{one condition in ORDER BY}
\State $g \gets [\ ]$ \Comment{order list}
\State $\mathcal{O} \gets \{\ \}$  \Comment{order dictionary}

\If{\Call{getChildCount}{$c_{OB}$} == 1}
    \State $ch \gets \Call{visit}{c_{OB}.child(0)}$
    \State $\mathcal{O}[ch] \gets$ "ASC" \Comment{default sorting}
\Else
    \For{$ch \in \Call{getChildren}{c_{OB}}$}
        \If{\Call{isTerminal}{$ch$}}
            \State $t \gets \Call{upper}{ch.text}$
        \Else \Comment{expression or variable}
            \State $t \gets \Call{visit}{ch}$
        \EndIf

        \If{$t$ in ["(", ")"]}
            \State continue \Comment{ignore any brackets}
        \EndIf

        \State $g \underset{app}{\gets} t$ \Comment{append to $g$}
    \EndFor
\EndIf

\If{$\Call{len}{g}$ == 2}
    \State $\mathcal{O}[g[1]] \gets g[0]$
\EndIf

\State $\mathcal{O}_{NS} \gets \{\ \}$ \Comment{$\mathcal{O}$ with namespaced node properties}
\For{$k \in \Call{keys}{\mathcal{O}}$}
    \State $k_{NS} \gets k$
    \If{$\neg \Call{keys}{\mathcal{G}_{WITH}}$ \textbf{and} $k \notin \mathcal{G}_{RETURN}$}
        \State $k_{NS} \gets \mathcal{G}_{PROPS}[k] \gets k$
    \ElsIf{$k \notin \mathcal{G}_{WITH}$}
        \State $\mathcal{G}_{WITH}[k] \gets \texttt{"}\mathcal{G}_{PROPS}[k]\texttt{ AS } k\texttt{"}$
    \EndIf
    \State $\mathcal{O}_{NS}[k_{NS}] \gets \mathcal{O}[k]$
\EndFor
\State $\mathcal{G}_{OB} \underset{upd}{\gets} \mathcal{O}_{NS}$ \Comment{update ORDER BY map}
\end{algorithmic}
\end{algorithm}

\subsubsection{Visit LIMIT}
\label{alg:visit_limit}
\begin{algorithm}[H]
\caption{visitLimitClause}
\begin{algorithmic}
\Require $val$ \Comment{LIMIT clause value}
    \State $\mathcal{G}_{LIMIT} \gets \Call{int}{val}$
\end{algorithmic}
\end{algorithm}

\subsubsection{Visit OFFSET}
\label{alg:visit_offset}
\begin{algorithm}[H]
\caption{visitOffsetClause}
\begin{algorithmic}
\Require $val$ \Comment{OFFSET clause value}
    \State $\mathcal{G}_{SKIP} \gets \Call{int}{val}$
\end{algorithmic}
\end{algorithm}

\subsection{Handling Functions}
\subsection{Visit Variable}
\label{alg:visit_var}
\begin{algorithm}[H]
\caption{visitVar}
\begin{algorithmic}
\Require $v$, $ctx$ \Comment{variable name and context}
\State $v \gets \Call{strip}{v, "?\$\ "}$
\If{$\Call{isInstance}{ctx, \texttt{aggregateContext}}$}
    \If{$v \in \mathcal{G}_{PROPS}$}
        \State $v \gets \mathcal{G}_{PROPS}[v]$
    \EndIf
\EndIf
\State $\mathcal{G}_\mathcal{V}[i] \gets v$
\State return " " + $v$
\end{algorithmic}
\end{algorithm}

\subsection{Visit Aggregate or BuiltIn function}
\label{alg:visit_aggregate}
\begin{algorithm}[H]
\caption{visitAggregate}
\begin{algorithmic}
\Require $\mathcal{E}$ ... expression, $\mathcal{A}$ ... alias (if exists)
\State $i \gets$ \Call{length}{$\mathcal{G}_{AGG}$}
\If{NOT \Call{Exists}{$\mathcal{A}$}}
    \State $\mathcal{A} \gets$ "agg\_\_\{i\}"
\EndIf
\State $\mathcal{G}_{WITH}[\mathcal{A}] \gets \mathcal{E}$ AS $\mathcal{A}$
\State $\mathcal{G}_{AGG}[\mathcal{A}] \gets \mathcal{E}$
\State return $\mathcal{A}$
\end{algorithmic}
\end{algorithm}

\newpage

\section{AST example}
\label{app:ast-example}
Given the parse tree in Figure \ref{fig:sparql-query-01-parse-tree} (from SPARQL query shown in Figure \ref{fig:visitor-overview}, resp. Listing \ref{lst:sparql-query-01}),  Visitor produces an abstract syntax tree ($\mathcal{G}$) shown in Listing \ref{lst:query-01-ast}.

\begin{lstlisting}[basicstyle=\footnotesize,caption={AST produced by Visitor by traversing parse tree in Figure \ref{fig:sparql-query-01-parse-tree}.},label=lst:query-01-ast]
{
  "vars": [
    "pet",
    "avgPersonAge",
    "x",
    "petName",
    "personAge"
  ],
  "iri": {},
  "nodes": {
    "x": {
      "label": "ROOT__Person"
    },
    "pet": {
      "label": "ROOT__Pet"
    }
  },
  "props": {
    "personAge": "x.person__age",
    "petName": "pet.pet__name"
  },
  "rels": [
    {
      "s": "x",
      "r": ":person__hasPet",
      "o": "pet",
      "optional": false,
      "inverse": false
    }
  ],
  "rel_types": [
    ":person__hasPet"
  ],
  "aggregates": {
    "avgPersonAge": "AVG(x.person__age)"
  },
  "WHERE": [
    "pet.pet__name CONTAINS 'b'"
  ],
  "WITH": {
    "avgPersonAge": "AVG(x.person__age) AS avgPersonAge",
    "petName": "pet.pet__name AS petName"
  },
  "WHERE_WITH": [
    [
      "avgPersonAge",
      ">",
      "30"
    ]
  ],
  "RETURN": [
    "petName, avgPersonAge"
  ],
  "ORDER BY": {
    "avgPersonAge": "DESC"
  },
  "LIMIT": 10,
  "OFFSET": 1
}

\end{lstlisting}

\onecolumn

\section{Interpreter implementation details}
\label{sec:interpreter-implementation}
Given an arbitrary AST ($\mathcal{G}$), Interpreter defines the builder classes below to construct respective Cypher clauses and concatenates them to produce the final Cypher query. For an example of $\mathcal{G}$, refer to Lst. \ref{lst:query-01-ast}.

\subsection{MATCHBuilder}
The MATCHBuilder class constructs the MATCH clause by iterating over the nodes ($\mathcal{G}_{\mathcal{N}}$) and rels ($\mathcal{G}_{\mathcal{R}}$) in the AST to generate Cypher MATCH statements.

\begin{lstlisting}[caption={MATCHBuilder Class implementation}, label=lst:match_builder]
class MATCHBuilder:
    def __init__(self, graph: dict):
        self.g = graph
        self.match = ""
        self.build()

    def build(self):
        nodes = self.g["nodes"]
        for entry in self.g["rels"]:
            s, rel_seq, o = entry["s"], entry["r"], entry["o"]
            s_node = f"({s}:{nodes[s]['label']})"
            o_node = f"({o}:{nodes[o]['label']})"
            rel_str = "-[]-".join([f"[{r}]" for r in rel_seq.split("/")])
            self.match += f"MATCH {s_node}-{rel_str}->{o_node}\n"

    def get(self):
        return self.match.strip()
\end{lstlisting}

\subsection{RETURNBuilder}
The RETURNBuilder class constructs the RETURN clause by utilising the $\mathcal{G}_{RETURN}$ field in the AST to determine which variables and expressions should be projected to the output.

\begin{lstlisting}[caption={RETURNBuilder Class implementation}, label=lst:return_builder]
class RETURNBuilder:
    def __init__(self, graph: dict):
        self.g = graph
        self.return_str = ""
        self.build()

    def build(self):
        term = self.g["RETURN"][0]
        terms = term.split(", ")
        self.return_str = "RETURN " + ", ".join(terms)

    def get(self):
        return self.return_str.strip()
\end{lstlisting}

\subsection{WHEREBuilder}
The WHEREBuilder class constructs the WHERE clause by converting conditions in the $\mathcal{G}_{Where}$ field of the AST into a Cypher-compatible format.

\begin{lstlisting}[caption={WHEREBuilder Class implementation}, label=lst:where_builder]
class WHEREBuilder:
    def __init__(self, graph: dict):
        self.g = graph
        self.term_list = []
        self.where = ""

        for term in self.g["WHERE"]:
            # print(term)
            self.add(term)

        self.build()

    def add(self, term):
        if isinstance(term, dict):
            print(NotImplementedError("Nested SELECT statements not supported yet."))
            return False

        if isinstance(term, list):
            # handle RelationalSetExpressions
            t = _construct_relational_set_expression(term)
        else:
            if "CONTAINS" in term:
                self.term_list.append(term)
                return True

            or_term = term.replace(",", "").replace("( ", "").replace(" )", 
                                                                "").split("||")
            and_or_terms = [orterm.strip().split("&&") for orterm in or_term]
            t = ""
            for subterm in and_or_terms:
                t += "(" + " AND ".join(subterm) + ")" + " OR "

        t = t.removesuffix(" OR ")
        self.term_list.append(t)
        return True

    def build(self):
        where_expr = []
        for term in self.term_list:
            for alias, agg in self.g["aggregates"].items():
                term = term.replace(alias, agg)
            where_expr.append(term)

        if where_expr:
            self.where = "WHERE " + " AND ".join(where_expr)

    def get(self):
        return self.where

    def clear(self):
        self.having_list = []
        self.wth, self.where = (None, None)
\end{lstlisting}

\subsection{WITHBuilder}
WITHBuilder gathers variables and expressions from the $\mathcal{G}_{WITH}$ field in the AST to facilitate aggregation, group variables and aliasing. 

\begin{lstlisting}[caption={WITHBuilder Class implementation}, label=lst:with_builder]
class WITHBuilder:
    def __init__(self, graph: dict):
        self.g = graph
        self.with_list = []
        self.wth = ""
        self.build()

    def build(self):
        wth_terms = ", ".join(self.g["WITH"].values())
        self.wth = f"WITH {wth_terms}"

    def get(self):
        return self.wth.strip()
\end{lstlisting}

\subsection{WWBuilder}
The WHERE clause after the first WITH is built using the $\mathcal{G}_{WhereWith}$ part of the AST in Lst. \ref{lst:ww_builder}. This serves in similar light as HAVING clause in SPARQL (see \ref{subsubsec:constraints}). 

\begin{lstlisting}[caption={WhereWithBuilder Class implementation}, label=lst:ww_builder]
class WWBuilder:
    def __init__(self, graph: dict):
        self.g = graph
        self.having_list = []
        self.where = ""

        for term in self.g["WHERE_WITH"]:
            self.add(term)

        self.build()

    def add(self, term):
    term = list(term)
        if isinstance(term, list):
            # handle RelationalSetExpressions
            t = _construct_relational_set_expression(term)
        else:
            or_term = term.replace(",", "").replace("( ", "").replace(" )", "").split("||")
            and_or_terms = [orterm.strip().split("&&") for orterm in or_term]
            t = ""
            for subterm in and_or_terms:
                t += "(" + " AND ".join(subterm) + ")" + " OR "

        t = t.removesuffix(" OR ")
        self.having_list.append(t)

    def build(self):
        where_expr = []
        if not self.having_list:
            logger.info("No HAVING clause found in the query.")
            return

        for term in self.having_list:
            where_expr.append(term)

        if where_expr:
            self.where = "WHERE " + " AND ".join(where_expr)

    def get(self):
        return f"{self.where}"
\end{lstlisting}

\subsection{SMBuilder}
The Solution Modifier builder class constructs the ORDER BY, LIMIT and SKIP cypher clauses, modifying the solution set of the query.

\begin{lstlisting}[caption={SMBuilder Class implementation}, label=lst:sm_builder]
class SMBuilder:
    def __init__(self, graph: dict):
        self.g = graph
        self.order_by = ""
        self.limit = ""
        self.skip = ""
        self.build()

    def build(self):
        if self.g["LIMIT"]:
            self.limit = f"LIMIT {self.g['LIMIT']}"
        if self.g["SKIP"]:
            self.skip = f"SKIP {self.g['SKIP']}"
        if self.g["ORDER BY"]:
            self.order_by = "ORDER BY " + ", ".join(
                [f"{term} {direction}" for term, direction 
                 in self.g["ORDER BY"].items()]
                )

    def get_all(self):
        return "\n".join([self.order_by, self.limit, self.skip]).strip()
\end{lstlisting}

\subsection{Assembling the Final Cypher Query}
The final Cypher query is assembled by concatenating the outputs of the builder classes in the correct order. Each builder contributes its respective part to the complete query as shown in Lst. \ref{lst:parse-ast-to-cypher} and eq. (\ref{eq:parse-ast-to-cypher}).

\begin{equation}
\label{eq:parse-ast-to-cypher}
\text{Cypher} = \text{MATCH} + \text{WHERE} + \text{WITH} + \text{WW} + \text{UNWIND} + \text{RETURN} + \text{SM}
\end{equation}

\onecolumn

\section{Error analysis}
\subsection{BSBM42}

\subsubsection{OptionalRelationship}
\label{subsubsec:err-bsbm-opt-rel}
The \textit{OptionalRelationship} SPARQL query (see \ref{lst:bsbm-optrel-sparql}) was created by \citet{zhao2023s2ctrans} to test situations, where OPTIONAL block contains relationships (type 4 pattern in \ref{tab:s2c-pattern-matching}). In this case, the Cypher query must contain OPTIONAL MATCH clause. Both S2CTrans (see Lst. \ref{lst:bsbm-optrel-s2ctrans}) and S2CLite (see Lst. \ref{lst:bsbm-optrel-s2clite}) parse the SPARQL query without major errors. The queries are almost identical, disregarding projection aliasing (which only S2CLite does) and unnecessary restatement of node labels. However, S2CLite fails at this query, as it produces wrong output row values when executed on BSBM. See execution output for the S2CTrans query in Lst. \ref{lst:bsbm-optrel-s2ctrans-out} and for S2CLite in Lst. \ref{lst:bsbm-optrel-s2clite-out} and note that S2CLite does not match to the execution output of original SPARQL query in Lst. \ref{lst:bsbm-optrel-sparql-out}.

\paragraph{Cause} of the discrepancy in outputs, is visible from the two parsed Cypher queries. S2CLite puts OPTIONAL MATCH in front of WHERE, while S2CTrans puts if after the WHERE statement. The original SPARQL query has OPTIONAL block before the FILTER, which would syntactically favour the S2CLite version. Contrary to our intuition, S2CTrans produces the correct results while swapping the original order of the equivalent clauses (first WHERE then OPTIONAL MATCH vs first OPTIONAL block then FILTER).

\begin{lstlisting}[caption={\textit{SPARQL query "OptionalRelationship.rq" from BSBM42 crafted by 
\citet{zhao2023s2ctrans}}}, label=lst:bsbm-optrel-sparql]
PREFIX bsbm-inst: <http://www4.wiwiss.fu-berlin.de/bizer/bsbm/v01/instances/>
PREFIX bsbm: <http://www4.wiwiss.fu-berlin.de/bizer/bsbm/v01/vocabulary/>
PREFIX rdfs: <http://www.w3.org/2000/01/rdf-schema#>
PREFIX rdf: <http://www.w3.org/1999/02/22-rdf-syntax-ns#>
PREFIX ele: <http://purl.org/dc/elements/1.1/>

SELECT distinct ?label WHERE{ 
	?review rdf:type bsbm:Review .
	?review bsbm:reviewFor ?product .
	?product rdf:type bsbm-inst:ProductType1. 
	?product bsbm:productPropertyNumeric1 ?pPN1 .
	?product rdfs:label ?label. 
	OPTIONAL {
		?product ele:publisher ?producer .
		?producer a bsbm:Producer .
	}
	FILTER(?pPN1 > 1000)
}
ORDER BY(?label) 
LIMIT 10
\end{lstlisting}

\begin{lstlisting}[caption={\textit{S2CTrans Cypher query parsed from SPARQL Lst. \ref{lst:bsbm-optrel-sparql} (OptionalRelationship).}}, label=lst:bsbm-optrel-s2ctrans]
MATCH (review:bsbm__Review)-[reviewFor:bsbm__reviewFor]->
(product:bsbm_inst__ProductType1)
WHERE product.bsbm__productPropertyNumeric1 > 1000
OPTIONAL MATCH (product)-[publisher:ele__publisher]->
(producer:bsbm__Producer)
RETURN DISTINCT product.rdfs__label
ORDER BY product.rdfs__label ASC
LIMIT 10    
\end{lstlisting}

\begin{lstlisting}[caption={\textit{S2CLite Cypher query parsed from SPARQL Lst. \ref{lst:bsbm-optrel-sparql} (OptionalRelationship).}}, label=lst:bsbm-optrel-s2clite]
MATCH (review:bsbm__Review)-[:bsbm__reviewFor]->
(product:bsbm_inst__ProductType1)
OPTIONAL MATCH (product:bsbm_inst__ProductType1)-[:ele__publisher]->
(producer:bsbm__Producer)
WHERE product.bsbm__productPropertyNumeric1 > 1000
RETURN DISTINCT product.rdfs__label AS label
ORDER BY product.rdfs__label ASC
LIMIT 10
\end{lstlisting}

\begin{lstlisting}[caption={\textit{Output rows of original SPARQL query in Lst. \ref{lst:bsbm-optrel-sparql} (OptionalRelationship).}}, label=lst:bsbm-optrel-sparql-out]
[{'label': 'abjured'},
{'label': 'absorbs'},
{'label': 'abysm'},
{'label': 'abysms'},
{'label': 'accommodator'}, 
{'label': 'acidness'},
{'label': 'acquirement'}, 
{'label': 'adorned'},
{'label': 'adsorptiveness'},
{'label': 'adulterously'}]
\end{lstlisting}

\begin{lstlisting}[caption={\textit{Output rows of S2CTrans Cypher query parsed from SPARQL Lst. \ref{lst:bsbm-optrel-sparql} (OptionalRelationship).}}, label=lst:bsbm-optrel-s2ctrans-out]
[{'label': 'abjured'},
{'label': 'absorbs'},
{'label': 'abysm'},
{'label': 'abysms'},
{'label': 'accommodator'}, 
{'label': 'acidness'},
{'label': 'acquirement'}, 
{'label': 'adorned'},
{'label': 'adsorptiveness'},
{'label': 'adulterously'}]
\end{lstlisting}

\begin{lstlisting}[caption={\textit{Output rows of S2CLite Cypher query parsed from SPARQL Lst. \ref{lst:bsbm-optrel-sparql} (OptionalRelationship).}}, label=lst:bsbm-optrel-s2clite-out]
[{'label': 'aardvark'},
{'label': 'ab areaways'},
{'label': 'abdicable'},
{'label': 'abjured'},
{'label': 'ablatively'},
{'label': 'abnegators'},
{'label': 'abnormality'},
{'label': 'abolishment'},
{'label': 'abolitionism'},
{'label': 'abominated'}]
\end{lstlisting}

\twocolumn

\section{Execution accuracy evaluation}
\label{sec:execution-accuracy-script}
This section describes the logic, rules, and limitations used to evaluate the equivalence of results between SPARQL and the parsed Cypher queries. We compare results by aligning the formatting of returned rows, given some general transformations and assumptions.

\subsection{General Assumptions}
We assume that the underlying knowledge graphs are correctly mapped to each other, meaning that entities, relationships, and properties align structurally and semantically. We use Neo4j neosemantics \cite{barrasa2021neosemantics} to load the serialized rdf graph (Turtle) files into the property graph, which adheres to Neo4j standards. Thus, we consider the databases to be valid and equivalent and any mismatch in results to stem from query execution or translation differences rather than graph inconsistencies. To evaluate equivalence, we compare the query results across three aspects:
\paragraph{Row Count (NUM\_RES):} Ensure the number of result rows matches. The total number of rows returned by both queries must be identical. If the row counts differ, the result is marked as a \texttt{NUM\_RES} mismatch error.
\paragraph{Values (VAL):} For corresponding rows, the values in all columns must match exactly. Any discrepancy in individual cell values results in a \texttt{VAL} mismatch error.
\paragraph{Execution error (N4j\_EXEC):} If SPARQL query is executed, but Cypher query throws an execution error. This includes queries that fail to execute due to syntax errors or execution errors.

\subsection{Special cases}
\paragraph{SP 0)} Handle rounding errors in floats. When comparing float values, we allow a small rounding error (1e-6).
\paragraph{SP 1)} Handle values 0 and null. For some aggregations, null value is represented as 0 in Neo4j, while in RDFlib it is always null. Thus we universally classify 0 and null as equivalent values.
\paragraph{SP 2)} Handle Non-deterministic ordering of results. This is necessary because of non-deterministic ordering between RDF and Neo4j is not consistent. Thus, we first sort the results by their values.
\paragraph{SP 3)} Handle NULL or Empty ([]) Results. Queries returning empty result sets are considered equivalent if both queries return no results. If only one query returns an empty list, but the other doesn't it is categorized as a \texttt{NUM\_RES} mismatch error.
\paragraph{SP 4)} Handle results with full nodes as returned bindings. As the representation of `node` is different between RDF graphs and property graphs, we compare node equivalency based on the URI of the node. If URIs of the returned nodes are equal between SPARQL and Cypher bindings, they are considered matching results.

\subsection{Comparison Process}
The comparison process is implemented using the following steps:
\begin{enumerate}
    \item \textbf{Query Execution:} Both SPARQL and Cypher queries are executed. 
    \item \textbf{Result Parsing:} Results are parsed and normalized into comparable formats.
    \item \textbf{Mismatches Categorization:} Any discrepancies are categorized into one of the predefined mismatch categories; \texttt{NUM\_RES} mismatch, \texttt{VAL} mismatch or \texttt{Neo4j\_EXEC} error.
\end{enumerate}

\subsection{Edge Cases and Limitations}
\begin{itemize}
    \item \textbf{Ignoring variable names mismatch:} The set of variable names (keys) in the results do not have to match exactly between SPARQL and Cypher.
    \item \textbf{SPARQL syntax errors:} Some of the SPARQL queries result in syntax error when executed on their respective RDF database. We exclude these queries from the comparison and remove them from the total count of queries.  
\end{itemize}

\end{document}